\documentclass[twocolumn]{template}

\usepackage[utf8]{inputenc}             
\usepackage[T1]{fontenc}                
\usepackage{url}                        
\usepackage{xurl}                       
\usepackage{booktabs}                   
\usepackage{multirow}
\usepackage{colortbl}
\usepackage{multicol}
\usepackage{placeins}
\usepackage{etoolbox}
\usepackage{amsfonts}                   
\usepackage{nicefrac}                   
\usepackage{microtype}                  
\usepackage[dvipsnames]{xcolor}         

\usepackage{latexsym}

\usepackage{graphicx}
\usepackage{float}
\usepackage{subcaption}
\usepackage{wrapfig}
\graphicspath{{figures/}}
\usepackage{cuted}
\usepackage{bm}

\usepackage{tabularx} 
\usepackage{ragged2e} 
\newcolumntype{L}{>{\RaggedRight\hangafter=1\hangindent=0em}X}

\usepackage{enumitem}

\usepackage{amsmath}
\usepackage{amssymb}
\usepackage{mathtools}
\usepackage{amsthm}

\setboolean{logo}{true}    

\usepackage[linesnumbered,ruled,vlined]{algorithm2e}

\hypersetup{
    colorlinks=true,
    linkcolor=red,
    citecolor=Cerulean,
    filecolor=magenta,      
    urlcolor=magenta,
}

\usepackage[capitalize,noabbrev]{cleveref}
\crefname{section}{§}{§§}
\Crefname{section}{§}{§§}

\usepackage{calligra}
\DeclareMathAlphabet{\mathcalligra}{T1}{calligra}{m}{n}

\usepackage{pifont}

\theoremstyle{plain}

\theoremstyle{definition}

\theoremstyle{remark}

\renewcommand{\paragraph}[1]{\vspace{1mm}\noindent\textbf{#1}}

\DeclareCaptionLabelFormat{cont}{#1~#2\alph{ContinuedFloat}}
\usepackage[most]{tcolorbox}

\tcbset{
  promptbox/.style={
    enhanced jigsaw,
    breakable,
    width=\columnwidth,
    colback=lightgray!20,
    colframe=Black,
    colbacktitle=NavyBlue,
    coltitle=white,
    fonttitle=\bfseries\small,
    fontupper=\footnotesize,
    boxrule=0.7pt,
    arc=1.5mm,
    outer arc=1.5mm,
    left=2mm,
    right=2mm,
    top=1.5mm,
    bottom=1.5mm,
    lefttitle=2mm,
    righttitle=2mm,
    toptitle=1mm,
    bottomtitle=1mm,
    before upper={\raggedright\setlength{\parindent}{0pt}},
    before skip=6pt,
    after skip=8pt
  }
}

\newtcolorbox{promptbox}[2][]{%
  promptbox,
  title={\parbox{\dimexpr\linewidth-4mm\relax}{\raggedright #2}},
  #1
}
\tcbset{
  takeawaybox/.style={
    top=10pt,
    colback=lightgray!20,
    colframe=Black,
    colbacktitle=BurntOrange,
    enhanced,
    center,
    attach boxed title to top center={yshift=-0.1in,xshift=0.0in},
    boxed title style={boxrule=0pt,colframe=white,},
  }
}
\newtcolorbox{takeawaybox}[2][]{takeawaybox, title=#2,#1}
\tcbset{
  observationbox/.style={
    top=10pt,
    colback=lightgray!20,
    colframe=Black,
    colbacktitle=YellowGreen,
    enhanced,
    center,
    attach boxed title to top center={yshift=-0.1in,xshift=0.0in},
    boxed title style={boxrule=0pt,colframe=white,},
  }
}
\newtcolorbox{observationbox}[2][]{observationbox, title=#2,#1}

\usepackage{xspace}

\newcommand{\blfootnote}[1]{%
  \begingroup
  \renewcommand{\thefootnote}{}%
  \footnotetext{#1}%
  \endgroup
}

\usepackage{CJK}

\title{H²SD: Hybrid Hindsight Self-Distillation}

\author[1,2]{Qiye Cai{*}}
\author[2,3]{Yichuan Ma{*}}
\author[1]{Weitao Ma}
\author[2,3]{Peiji Li}
\author[2]{Yongkang Chen}
\author[2]{Yicheng Zou}
\author[2,4]{Linyang Li\textsuperscript{\textdagger}}
\author[1]{Xiaocheng Feng\textsuperscript{\textdagger}}
\author[1]{Bing Qin\textsuperscript{\textdagger}}

\affil[1]{Harbin Institute of Technology}
\affil[2]{Shanghai Artificial Intelligence Laboratory}
\affil[3]{Fudan University}
\affil[4]{The Chinese University of Hong Kong}

\begin{abstract}
Reinforcement learning with verifiable rewards (RLVR) provides reliable outcome supervision for language model reasoning, but scalar trajectory rewards offer limited guidance for token-level credit assignment. Existing self-distillation methods incorporate privileged teacher information through fixed strategies: direct distribution matching may destabilize successful behavior, while magnitude modulation provides limited corrective guidance after failure. We propose Hybrid Hindsight Self-Distillation (H$^2$SD), which adapts both teacher context and update strategy to trajectory correctness. For successful trajectories, the teacher conditions on the verified student response and a rephrasing instruction to rescore the original tokens. These scores modulate reward-derived updates while preserving advantage signs. For failed trajectories, a verified reference hint guides corrective supervision through reverse KL distillation. Across Sudoku, Calcudoku, Arrow Maze, and mathematical reasoning with two Qwen3 models, H$^2$SD achieves an overall average accuracy of 43.33\%, outperforming the strongest baseline by 11.43 percentage points while maintaining stable optimization.
Code is available at \url{https://github.com/kle1n2/H2SD}.
\end{abstract}

\begin{document}

\maketitle
\blfootnote{%
\textsuperscript{*} Equal contribution.
\quad
$\dagger$ Corresponding authors: Linyang Li(lilinyang@pjlab.org.cn), Xiaocheng Feng
(xcfeng@ir.hit.edu.cn), Bing Qin (qinb@ir.hit.edu.cn) 
}

\section{Introduction}

Since the release of DeepSeek-R1~\citep{dsr1}, reinforcement learning with verifiable rewards (RLVR), represented by GRPO~\citep{grpo}, has been widely adopted to enhance the reasoning abilities of large language models (LLMs).
Meanwhile, the field of agentic RL has also rapidly progressed, where models interact with environments and improve through feedback and reward signals~\citep{toolrl,torl}. By carefully designing environmental feedback and reward functions, general LLMs have achieved remarkable improvements in tool use, open-ended environment interaction, and other agentic tasks~\citep{kimi2.5,glm5}.

\begin{figure}[t]
    \centering
    \includegraphics[width=0.9\linewidth]{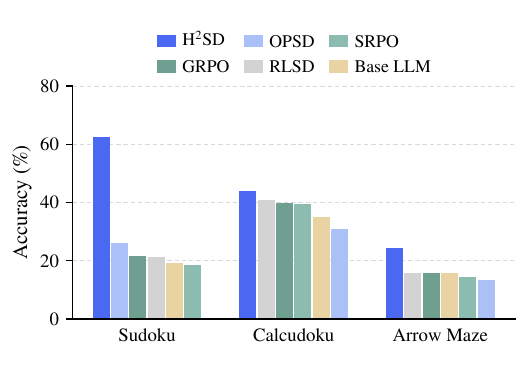}
    \caption{Overall performance comparison of Qwen3-30B-A3B-Instruct on representative logical reasoning benchmarks. H$^2$SD consistently delivers strong performance across logic puzzle tasks.
    }
    \label{fig:h2sd_intro}
\end{figure}

Despite these advances, RLVR algorithms usually obtain a single feedback signal from the environment and use it as a scalar outcome reward for optimization, which creates a bottleneck in credit assignment, since the reward provides limited information about which token is responsible for success or failure. On-policy distillation (OPD) offers a possible solution to this problem~\citep{opd-thinkingmachine,opdllm}. By introducing a stronger model as the teacher model and using the teacher’s token-level logits to supervise trajectories sampled by the student model, OPD provides denser supervision over the entire response.

However, OPD relies on a stronger teacher model, and the teacher and student models often need to share the same tokenizer. These requirements restrict its applicability in broader settings. To address this issue, on-policy self-distillation (OPSD) proposes an alternative paradigm~\citep{sdpo,opsd,sdft}. In OPSD, the same model is used as both the teacher and the student. The teacher model is provided with privileged information $r$, and, conditioned on this information, produces token-level supervision on trajectories sampled by the student. In this way, OPSD attempts to approximate the effect of OPD without actually requiring an external stronger teacher model.

Although OPSD is effective, directly matching a teacher distribution conditioned on privileged information may cause training instability and privileged information leakage, as observed in RLVR with Self Distillation (RLSD; \citealp{rlsd}). RLSD mitigates these risks by using the teacher signal only to modulate token-level update magnitudes. SRPO further routes successful samples to GRPO and failed samples to self-distillation~\citep{srpo}. 

However, this improved stability comes at the cost of weaker corrective supervision. Unlike OPSD, which transfers the teacher's fine-grained distributional preferences, RLSD mainly reweights the updates of tokens already generated by the student. Consequently, when the student follows an incorrect reasoning path, magnitude modulation alone may be insufficient to correct its underlying generation distribution.

To resolve this tension, we propose Hybrid Hindsight Self-Distillation (H$^2$SD), which assigns different roles to the teacher according to trajectory correctness. For successful trajectories, the reward already provides a reliable optimization direction. The teacher receives the student response confirmed as correct together with a rephrasing instruction, and its probabilities on the original response tokens are used only to modulate the update magnitudes in RLSD. For failed trajectories, we condition the teacher on a reference hint and minimize the reverse KL divergence from the student to the teacher. This preserves the direction determined by the reward for successful trajectories while providing explicit distributional correction when the sampled reasoning fails.

The quality of the teacher signal also depends on the privileged information it receives. We use a stronger language model to generate a reference hint for each problem, containing key intermediate reasoning steps and a final answer confirmed by the verifier. OPSD, RLSD, SRPO, and H$^2$SD have access to the same reference hints, controlling for the quality and availability of reference information.

With these designs, H$^2$SD achieves strong overall performance on logic puzzles and mathematical reasoning with both Qwen3-30B-A3B-Instruct and Qwen3-14B. These results support the effectiveness of adapting teacher supervision to trajectory correctness: refining credit assignment for individual tokens in successful trajectories and using a teacher distribution conditioned on reference hints to correct failed trajectories. Our main contributions are summarized as follows:

\begin{itemize}
    \item We identify a tension between correcting the policy distribution and maintaining stable credit assignment in on-policy self-distillation. We further show that the effectiveness of these supervision strategies depends on whether the student trajectory is correct.

    \item We propose H$^2$SD, a self-distillation framework that adapts teacher context and update strategy according to trajectory correctness. For successful trajectories, the teacher rescores the original student response under a rephrasing instruction to guide magnitude modulation. For failed trajectories, the student learns from a teacher distribution conditioned on reference hints through reverse KL minimization.

    \item We conduct experiments on Sudoku, Calcudoku, Arrow Maze, and mathematical reasoning across two model scales. H$^2$SD achieves an overall average accuracy of 43.33\% across task groups and model scales, outperforming the strongest baseline by 11.43 percentage points while maintaining stable optimization.
\end{itemize}

\section{Related Work}

\noindent\textbf{Reinforcement Learning with Verifiable Rewards.}
RLVR improves LLM reasoning through automatically verifiable rewards. GRPO~\citep{grpo} estimates group-relative advantages without a value model, while DAPO and GSPO~\citep{dapo,gspo} improve training stability and credit assignment. GOLF~\citep{golf} aggregates external critiques and attempts within each group into natural language refinements to guide exploration when rewards are sparse.

\noindent\textbf{On-Policy Distillation.}
On-policy distillation~\citep{opd-thinkingmachine,opdllm,minillm} mitigates the distribution shift problem of off-policy distillation by allowing the student model to generate its own trajectories and using a teacher model to evaluate these on-policy rollouts, providing dense token-level supervision. However, such methods usually rely on an additional strong teacher, and logit-level distillation often requires the teacher and student to have compatible token spaces, which limits the choice of teacher models.

\noindent\textbf{On-Policy Self-Distillation with Privileged Information.}
To reduce reliance on external teachers, recent studies have explored building self-teachers with privileged information. OPSD~\citep{opsd} allows the same model to serve as both student and teacher, where the teacher branch receives privileged ground-truth solutions and guides the student through token-level distribution matching. SDPO~\citep{sdpo} further broadens privileged information by incorporating environment feedback or correct rollouts as teacher signals. SD-Zero~\citep{zero} introduces a reviser to refine responses based on model outputs and rewards. Other methods, such as OPCD~\citep{opcd} and GATES~\citep{gates}, extend privileged information beyond final answers to include richer auxiliary signals such as context, experience, and evidence.

\noindent\textbf{Stable Self-Distillation for RLVR.}
Directly fitting a privileged teacher may introduce information leakage and training instability. RLSD~\citep{rlsd} addresses this issue by letting the environment reward determine the update direction while using the teacher only to modulate token-level magnitudes. Related methods extend this reward-grounded formulation through entropy-aware confidence gating~\citep{egrsd} and contrastive evidence reweighting~\citep{cepo}, while SRPO~\citep{srpo} routes successful trajectories to GRPO and failed ones to self-distillation.

\section{Preliminaries}

Before introducing the proposed H$^2$SD method, we first describe the implementation strategies of GRPO, OPSD, and RLSD.   Throughout this section, $x$ denotes a problem and $y=(y_1,\ldots,y_T)$ denotes a model response.

\begin{figure*}[t]
    \centering
    \includegraphics[width=0.9\linewidth]{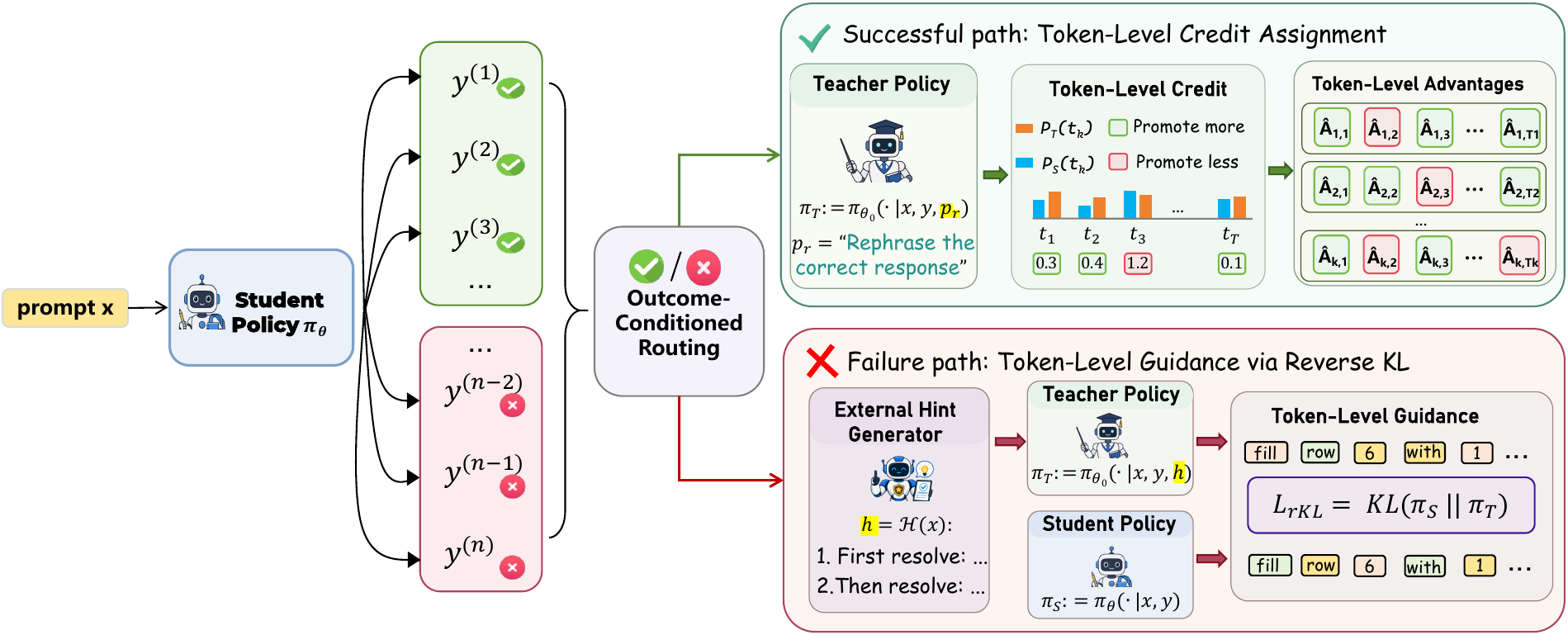}
    \caption{Overview of H$^2$SD. The framework routes student-generated trajectories according to their correctness: successful trajectories receive fine-grained token-level credit assignment, while failed trajectories are corrected through hint-conditioned distributional guidance.}
    \label{fig:h2sd}
\end{figure*}

\subsection{GRPO}

Given a question $x$, GRPO samples $G$ responses $\{y^{(i)}\}_{i=1}^{G}$ from the old policy, $y^{(i)}=(y_1^{(i)},\ldots,y_{T_i}^{(i)})\sim\pi_{\theta_{\mathrm{old}}}(\cdot\mid x)$, and a reward function assigns a scalar reward $R(x,y^{(i)})$ to each. GRPO then computes a relative advantage within the current group:
\par
\begin{equation}
A_i=\frac{R(x,y^{(i)})-\mu}{\sigma},
\end{equation}
where $\mu$ and $\sigma$ denote the mean and standard deviation of the rewards within the group. With the probability ratio $\rho_{i,t}(\theta)=\pi_\theta(y_t^{(i)}\mid x,y_{<t}^{(i)})/\pi_{\theta_{\mathrm{old}}}(y_t^{(i)}\mid x,y_{<t}^{(i)})$ and the token-level clipped surrogate $\ell_{i,t}(\theta)=\min\!\big(\rho_{i,t}(\theta)A_i,\operatorname{clip}(\rho_{i,t}(\theta),1-\epsilon,1+\epsilon)A_i\big)$, the GRPO objective is
\begin{equation}
\begin{aligned}
\mathcal{J}_{\mathrm{GRPO}}(\theta)
&=\frac{1}{G}\sum_{i=1}^{G}\frac{1}{T_i}\sum_{t=1}^{T_i}
\bigl[\ell_{i,t}(\theta)\\
&\qquad-\beta D_{\mathrm{KL}}(\pi_\theta\parallel\pi_{\mathrm{ref}})\bigr].
\end{aligned}
\end{equation}
All tokens within the same response therefore receive the same update magnitude, determined solely by the response-level advantage $A_i$. When trajectories are routed by correctness, $\mu$ and $\sigma$ are still computed over all $G$ responses; groups whose responses are all correct or all incorrect receive no policy-gradient signal.

\subsection{On-Policy Distillation and OPSD}

OPSD constructs a self-teacher by conditioning the same model on privileged information $r$. For a student trajectory $y\sim\pi_\theta(\cdot\mid x)$, we denote the student and teacher distributions at position $t$ as
\begin{equation}
\begin{aligned}
p_{t,\theta}^{\mathrm{T}}(\cdot)
&=\operatorname{sg}\bigl[\pi_\theta(\cdot\mid x,r,y_{<t})\bigr],\\
p_{t,\theta}^{\mathrm{S}}(\cdot)
&=\pi_\theta(\cdot\mid x,y_{<t}).
\end{aligned}
\end{equation}
OPSD directly matches the student distribution to this privileged teacher distribution, providing token-level supervision without an external teacher.

\subsection{RLSD}

Rather than directly matching the privileged teacher distribution, RLSD uses it to modulate the magnitude of the reward-based update. For a sampled token $y_t$, it computes
\begin{equation}
\Delta_t
=
\operatorname{sg}
\left(
\log P_T(y_t)
-
\log P_S(y_t)
\right),
\end{equation}

where $P^T(y_t)$ and $P^S(y_t)$ are the teacher and student probabilities of the sampled token. The detached token weight is $w_t=\exp(\operatorname{sign}(A)\Delta_t)$, where $A$ denotes the trajectory-level advantage. The resulting token-level advantage is

\begin{equation}
\begin{aligned}
\widehat{A}_{t}
&=A\bigl[(1-\lambda)\\
&\quad+\lambda\,\operatorname{clip}(w_t,1-\epsilon_w,1+\epsilon_w)\bigr],
\end{aligned}
\label{eq:magnitude}
\end{equation}
Thus, the teacher changes the magnitude of the reward derived update while preserving its sign.

\section{\texorpdfstring{H$^2$SD}{H²SD}: Hybrid Hindsight Self-Distillation}

\subsection{Hint Design}

RLSD typically derives privileged information from the ground-truth answer, whereas SDPO uses verified correct responses sampled during training. For complex reasoning tasks, ground-truth answers and rule-based feedback may provide limited process guidance, while verified responses can be scarce when the current policy has low accuracy.

To provide more informative supervision, we introduce a stronger LLM as a hint generator. Given a problem \(x\), the hint generator produces a natural-language hint \(h\) that provides useful information for solving the problem:
\begin{equation}
h = \mathcal{H}(x).
\end{equation}
The generated hint is used as privileged information that is available only during training to construct a self-teacher. The stronger model does not directly serve as the teacher and does not need to share the same vocabulary with the student model. It is only responsible for generating the natural-language hint. The actual teacher distribution is produced by a frozen copy $\theta_0$ of the initial policy conditioned on the hint:
\begin{equation}
\begin{aligned}
\pi_T(\cdot \mid x,h,y_{<t})
&=
\pi_{\theta_0}(\cdot \mid x,h,y_{<t}),\\
\pi_S(\cdot \mid x,y_{<t})
&=
\pi_\theta(\cdot \mid x,y_{<t}).
\end{aligned}
\end{equation}
Here $\theta_0$ is never updated or synchronized with the student, so the teacher shares the student's architecture and tokenizer and requires no external model during optimization. Because the teacher is used only to score tokens the student has already generated under hindsight context, it does not need to be stronger than the current policy.

\begin{table*}[t]
\centering
\begingroup
\setlength{\tabcolsep}{3pt}
\renewcommand{\arraystretch}{1.12}
\resizebox{\textwidth}{!}{%
\begin{tabular}{@{}l|ccc|ccc|cccccc|cccc|c@{}}
\toprule
\multirow{3}{*}{Method}
& \multicolumn{12}{c|}{Logic Puzzles}
& \multicolumn{4}{c|}{Math}
& \multirow{3}{*}{\shortstack{Overall\\Avg.}} \\
\cmidrule(lr){2-13}\cmidrule(lr){14-17}
& \multicolumn{3}{c|}{Sudoku}
& \multicolumn{3}{c|}{Calcudoku}
& \multicolumn{6}{c|}{Arrow Maze}
& \multirow{2}{*}{AIME 25}
& \multirow{2}{*}{AIME 26}
& \multirow{2}{*}{Minerva Math}
& \multirow{2}{*}{Avg.} & \\
\cmidrule(lr){2-4}\cmidrule(lr){5-7}\cmidrule(lr){8-13}
& 6$\times$6 & 8$\times$8 & Avg.
& 5$\times$5 & 6$\times$6 & Avg.
& 6$\times$6 & 7$\times$7 & 8$\times$8 & 9$\times$9 & 10$\times$10 & Avg.
& & & & & \\
\midrule
\multicolumn{18}{c}{Qwen3-30B-A3B-Instruct} \\
\midrule
Base LLM
& 24.50 & 14.00 & 19.25
& 59.00 & 11.33 & 35.17
& 32.00 & 17.50 & 16.00 & 6.00 & 6.50 & 15.60
& 63.13 & 76.04 & 45.89 & 61.69 & 32.93 \\
GRPO
& 26.25 & \underline{16.75} & 21.50
& \underline{66.33} & 13.67 & 40.00
& \underline{34.50} & 16.00 & 14.00 & \underline{7.50} & 6.50 & 15.70
& \textbf{75.63} & \underline{83.96} & 47.96 & \underline{69.18} & \underline{36.60} \\
RLSD
& 27.25 & 15.00 & 21.13
& 66.33 & \underline{15.33} & \underline{40.83}
& 34.00 & 17.00 & 15.00 & 7.00 & 6.00 & \underline{15.80}
& 69.38 & 79.38 & 47.31 & 65.36 & 35.78 \\
OPSD
& \underline{35.25} & 16.75 & \underline{26.00}
& 52.00 & 9.67 & 30.83
& 29.50 & 15.00 & 11.00 & 3.50 & \underline{7.00} & 13.20
& 64.17 & 77.29 & \underline{48.60} & 63.35 & 33.35 \\
SRPO
& 23.25 & 14.00 & 18.63
& 63.00 & \textbf{16.00} & 39.50
& 29.50 & \underline{18.50} & \underline{17.00} & 4.50 & 3.00 & 14.50
& 71.04 & 78.54 & 47.63 & 65.74 & 34.59 \\
\textbf{H$^2$SD}
& \textbf{76.50} & \textbf{48.50} & \textbf{62.50}
& \textbf{73.33} & 14.67 & \textbf{44.00}
& \textbf{52.00} & \textbf{29.50} & \textbf{21.00} & \textbf{9.00} & \textbf{9.50} & \textbf{24.20}
& \underline{73.75} & \textbf{87.08} & \textbf{48.78} & \textbf{69.87} & \textbf{50.14} \\
\midrule
\multicolumn{18}{c}{Qwen3-14B} \\
\midrule
Base LLM
& 12.00 & 0.00 & 6.00
& 29.67 & 1.67 & 15.67
& 0.00 & 0.00 & 0.00 & 0.00 & 0.00 & 0.00
& 69.38 & 66.04 & 51.72 & 62.38 & 21.01 \\
GRPO
& \underline{23.75} & 8.50 & 16.13
& \textbf{38.67} & \underline{2.00} & \textbf{20.33}
& 1.50 & 1.00 & 0.00 & 0.50 & 1.00 & 0.80
& \textbf{83.96} & 77.29 & 53.47 & \underline{71.57} & \underline{27.21} \\
RLSD
& 18.25 & 2.25 & 10.25
& 35.00 & 1.67 & 18.33
& 2.00 & 0.00 & 1.00 & 0.00 & 0.50 & 0.70
& 79.38 & \underline{80.63} & 53.61 & 71.21 & 25.12 \\
OPSD
& 7.25 & \underline{28.25} & \underline{17.75}
& 1.67 & 0.00 & 0.83
& \underline{21.50} & \underline{16.00} & \underline{13.00} & \underline{4.50} & \underline{3.50} & \underline{11.70}
& 70.63 & 72.71 & 53.79 & 65.71 & 24.00 \\
SRPO
& 15.75 & 6.25 & 11.00
& 31.33 & 1.33 & 16.33
& 3.50 & 1.00 & 1.50 & 0.50 & 1.00 & 1.50
& 76.04 & 76.04 & \underline{54.62} & 68.90 & 24.43 \\
\textbf{H$^2$SD}
& \textbf{27.00} & \textbf{36.50} & \textbf{31.75}
& \underline{36.67} & \textbf{3.33} & \underline{20.00}
& \textbf{51.50} & \textbf{24.00} & \textbf{17.50} & \textbf{7.50} & \textbf{9.50} & \textbf{22.00}
& \underline{81.04} & \textbf{80.83} & \textbf{55.03} & \textbf{72.30} & \textbf{36.51} \\
\bottomrule
\end{tabular}%
}
\endgroup
\caption{Validation accuracy (\%) of two Qwen3 models on logic puzzles (1 sample per problem) and math benchmarks (16 samples per problem). Overall Avg. is the macro-average across four task families. Bold/underlined values indicate the best/second-best results, including ties.}
\label{tab:main_results}
\end{table*}

\subsection{Hybrid Self-Distillation}

A straightforward approach is to directly match the student to a hint-conditioned teacher, as in OPSD. Although this provides process-level supervision, full distribution matching may expose privileged answer information at every token and cause information leakage. RLSD avoids direct matching by rescaling reward-directed updates, but cannot suggest alternatives after failure. Original SRPO partly addresses this limitation by routing failed trajectories to entropy-weighted SDPO, but its sibling solution may be unavailable on difficult problems and offers less explicit corrective guidance than a verified reference hint. For successful trajectories, SRPO still uses vanilla GRPO, whose trajectory-level rewards provide only coarse token-level credit assignment.

We therefore propose Hybrid Hindsight Self-Distillation (H$^2$SD), which selects the distillation strategy according to trajectory correctness. For successful trajectories, the positive reward already provides a reliable optimization direction. The teacher receives the student response verified as correct together with an instruction to rephrase it, and its probabilities on the original response tokens are used to compute the magnitude weights in RLSD. The teacher does not generate a rewritten response; the instruction only changes how it scores the original token sequence. This refines credit assignment across tokens without directly matching the teacher distribution or changing the direction determined by the reward.

For failed trajectories, the sampled reasoning path is unreliable. Although a negative reward discourages the sampled actions, it provides limited guidance on how they should be corrected. We therefore use the teacher distribution conditioned on the hint as corrective guidance and minimize the reverse KL divergence from the student to the teacher. Compared with the forward KL used by the OPSD baseline, reverse KL emphasizes suppressing student probability mass that is inconsistent with the hint-conditioned teacher rather than covering the teacher's full distribution, making it better suited to correcting erroneous reasoning after failure:
\begin{equation}
\mathcal{L}_{\mathrm{RKL}}
=
\frac{1}{T}\sum_{t=1}^{T}
D_{\mathrm{KL}}
\left(
p^{S}_{t,\theta}
\,\|\,
\operatorname{sg}[p^{T,h}_{t,\theta_0}]
\right),
\end{equation}
where $p^{S}_{t,\theta}=\pi_\theta(\cdot\mid x,y_{<t})$ and $p^{T,h}_{t,\theta_0}=\pi_{\theta_0}(\cdot\mid x,h,y_{<t})$. The operator $\operatorname{sg}[\cdot]$ blocks gradients through the teacher distribution.

Let $m=\mathbf{1}[R(x,y)=1]$ denote trajectory correctness. The overall objective is
\begin{equation}
\begin{aligned}
\mathcal{L}_{\mathrm{H^2SD}}
&=\mathbb{E}_{x,y}\bigl[m\mathcal{L}_{\mathrm{RLSD}}\\
&\quad+\gamma(1-m)\mathcal{L}_{\mathrm{RKL}}\bigr],
\end{aligned}
\end{equation}
where $\mathcal{L}_{\mathrm{RLSD}}$ is the loss form of the RLSD objective computed using the teacher prompted to rephrase, and $\gamma$ controls the strength of corrective distillation on failed trajectories. Thus, H$^2$SD refines credit assignment for successful trajectories and provides directional correction for failed trajectories.

\section{Experiments}

\subsection{Models and Benchmarks}
We evaluate two base models: Qwen3-30B-A3B-Instruct-2507 (denoted Qwen3-30B-A3B-Instruct) and Qwen3-14B on logic puzzle tasks and math benchmarks. The logic puzzle benchmarks cover three task families: Sudoku at sizes 6$\times$6 and 8$\times$8, Calcudoku at sizes 5$\times$5 and 6$\times$6, and Arrow Maze at sizes 6$\times$6 through 10$\times$10. Math benchmarks include AIME 25, AIME 26, and Minerva Math.

Specifically, Sudoku requires filling 6$\times$6 or 8$\times$8 grids without repeated digits in any row, column, or sub-grid. Arrow Maze assigns directions to cells so that each numbered ray has the required length and all cells are covered. Calcudoku combines row and column uniqueness with arithmetic constraints over predefined regions. Compared with widely reused mathematical datasets, these procedurally generated tasks reduce the risk of prior instance exposure and evaluate reasoning on newly generated problem instances.

\subsection{Implementation Details}
For each training problem, Kimi-K2.6 generates a reference hint. In order to fairly compare the various methods, for the baseline methods that use privileged information (OPSD, RLSD, SRPO) and H$^2$SD, we provide all methods that use privileged information with the same reference hints. Appendix~\ref{sec:method-specific-hyperparameters} provides more detailed configurations and parameters for each method.

For logic puzzles, we evaluate one response per problem, while for math benchmarks, we sample 16 responses per problem. We report the mean verifier acceptance rate, with family averages giving equal weight to the
reported sizes or benchmarks. 

We implement all methods in VERL with SGLang. At our model scales and sequence lengths, full-vocabulary distillation for OPSD incurred prohibitive memory costs. We therefore retain exact probabilities for the student's top-100 tokens and merge the remaining mass into a tail bucket. Training uses eight nodes, each with eight NVIDIA H200 140GB GPUs.

\subsection{Main Results}

\begin{figure}[t]
    \centering
    \includegraphics[width=\linewidth]
    {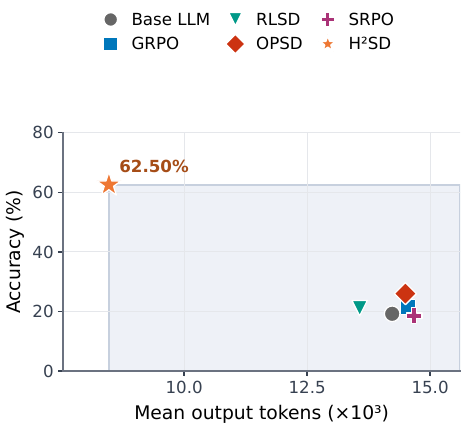}
    \caption{Accuracy--generation cost trade-off on Sudoku. Points toward the upper-left indicate better efficiency. H$^2$SD achieves the highest accuracy with the fewest generated tokens.}
    \label{fig:sudoku_length_accuracy}
\end{figure}

Table~\ref{tab:main_results} shows that H$^2$SD achieves the strongest overall performance across both model scales. On Qwen3-30B-A3B-Instruct and Qwen3-14B, H$^2$SD obtains overall average accuracies of 50.14\% and 36.51\%, outperforming GRPO, the strongest overall baseline, by 13.54 and 9.30 percentage points, respectively. Averaged across both models, H$^2$SD achieves 43.33\%, exceeding the strongest baseline by 11.43 percentage points. Since OPSD, RLSD, SRPO, and H$^2$SD use identical privileged hints, access to richer privileged information alone cannot explain these gains. Moreover, although SRPO similarly routes trajectories according to correctness, H$^2$SD consistently outperforms it overall, suggesting that how teacher supervision is incorporated matters beyond correctness-aware routing alone.

Across logic puzzles, H$^2$SD achieves particularly strong improvements on Sudoku and Arrow Maze. On Sudoku, it reaches 62.50\% and 31.75\% accuracy for the two models, exceeding the strongest task-specific baselines by 36.50 and 14.00 percentage points, respectively. On Arrow Maze, H$^2$SD improves upon the strongest baselines by 8.40 and 10.30 points. Improvements on Calcudoku are more modest, with H$^2$SD achieving the highest accuracy on the larger model while remaining competitive on the smaller one. The gains also extend to mathematical reasoning, where H$^2$SD achieves the highest math averages of 69.87\% and 72.30\% for the two models. Although these improvements are smaller than those observed on logic puzzles, they indicate that the benefits of H$^2$SD are not restricted to a single reasoning task or model scale.

Figure~\ref{fig:sudoku_length_accuracy} further shows that the gains of H$^2$SD do not rely on longer responses. It achieves the highest accuracy with the lowest average generation cost in all tasks (Appendix~\ref{app:length_analysis}), indicating that the performance gains primarily arise from more effective use of the available learning signals rather than increased response length.

\begin{figure}[t]
    \centering
    \includegraphics[width=\linewidth]{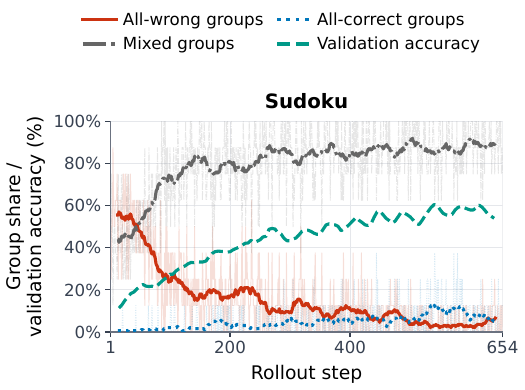}
    \caption{Training-group composition under H$^2$SD on Sudoku, averaged equally across grid sizes at shared rollout steps.}
    \label{fig:group_routing}
\end{figure}

\begin{figure*}[t]
    \centering
    \includegraphics[width=0.9\linewidth]{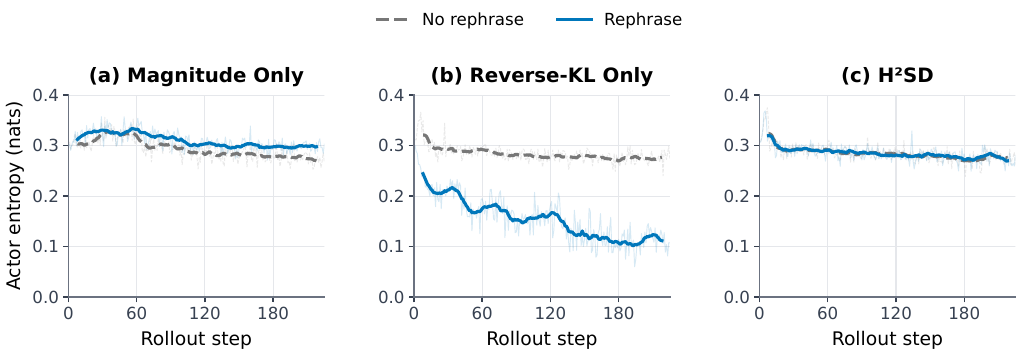}
    \caption{Actor entropy with and without successful-trajectory rephrasing on Sudoku, averaged equally across grid sizes at shared rollout steps. Light curves show recorded values; dark curves are 15-step centered means. Rephrasing coincides with a sharper entropy decrease under Reverse-KL Only.}
    \label{fig:routing_entropy}
\end{figure*}

\subsection{How Should Updates Be Assigned by Outcome?}

Each ablation varies only the studied factor while keeping the model (Qwen3-30B-A3B-Instruct), training data, rollout settings, and optimization hyperparameters fixed.
\begin{table}[t]
\centering
\small
\setlength{\tabcolsep}{3pt}
\begin{tabular}{@{}lccc@{}}
\toprule
\textbf{Update strategy} & \textbf{Sudoku} & \textbf{Calcudoku} & \textbf{Arrow Maze} \\
\midrule
Magnitude Only & 55.00 & 43.17 & 15.60 \\
Reverse-KL Only & 42.50 & 28.67 & 18.50 \\
Reversed Routing & 13.38 & 2.83 & 6.40 \\
GRPO on Success & 60.75 & 40.00 & 20.20 \\
H$^2$SD & \textbf{62.50} & \textbf{44.00} & \textbf{24.20} \\
\bottomrule
\end{tabular}
\caption{Average validation accuracy (\%) under different routing strategies. Magnitude Only and Reverse-KL Only apply one update to all trajectories,
while Reversed Routing swaps H$^2$SD's success/failure assignments. GRPO on Success removes only successful-trajectory teacher weighting.}
\label{tab:routing_ablation}
\end{table}

Table~\ref{tab:routing_ablation} compares alternative routing strategies across all three task families. GRPO on Success removes teacher-based token reweighting from successful trajectories and instead applies the standard GRPO update, while retaining the same group-relative advantages, loss normalization, and hint-conditioned reverse KL updates on failed trajectories. H$^2$SD outperforms both uniform strategies, while Reversed Routing falls below the Base LLM in every setting. These results support the chosen assignment within the tested training recipe.

Figure~\ref{fig:group_routing} illustrates how the rollout composition evolves together with validation performance. At initialization, 55.5\% of Sudoku rollout groups contain no correct response, causing the group-relative advantage to vanish for most prompts. Under H$^2$SD, the all-wrong fraction falls below 20\%
within 100 rollout steps, while validation accuracy steadily
increases. This pattern is consistent with the reverse KL branch providing useful supervision in the early all-wrong regime, after which successful trajectories become sufficiently frequent for magnitude modulation to contribute more often.

\begin{figure*}[!tp]
\centering
\includegraphics[width=0.9\textwidth]{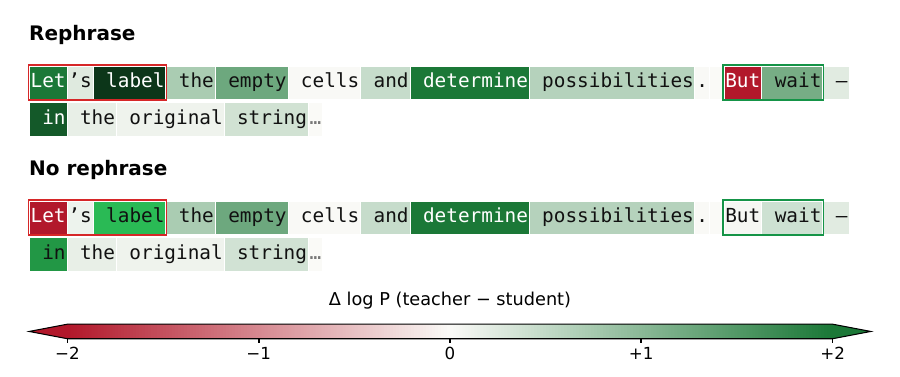}
\caption{An example of rephrasing effects on Sudoku tasks. Red and green outlines mark key deductions and redundant checks, respectively.}
\label{fig:teacher_student_schematic}
\end{figure*}

\subsection{Which Privileged Context Is Useful?}

We keep the H$^2$SD training recipe fixed and compare four types of privileged information: (1) \textbf{Programmatic feedback}, which provides fine-grained error feedback generated by a task-specific verifier based on the differences between the student response and the ground-truth solution; (2) \textbf{Ground truth}, which contains only the correct final answer without any intermediate reasoning; (3) \textbf{Sibling solution}, which uses a verified correct response sampled from multiple rollouts of the same problem as teacher guidance; and (4) \textbf{Hint}, which is a reference solution generated by an external strong model and contains both key intermediate reasoning steps and the final answer. In the Sibling solution condition, we fall back to a GRPO update when no verified correct sibling response is available.

As shown in Table~\ref{tab:privileged_context}, programmatic feedback and ground truth give modest gains on Sudoku and Calcudoku, but programmatic feedback reduces Arrow Maze accuracy. Sibling solutions improve Sudoku and Arrow Maze, while reducing Calcudoku accuracy below the Base LLM. Hints perform best in each column, supporting richer process-level context in the tested training recipe.

\begin{table}[t]
\centering
\small
\setlength{\tabcolsep}{1.8pt}
\begin{tabular}{@{}lccc@{}}
\toprule
\textbf{Context} & \textbf{Sudoku} & \textbf{Calcudoku} & \textbf{Arrow Maze} \\
\midrule
Base LLM & 19.25 & 35.17 & 15.60 \\
Programmatic feedback & 19.50 & 39.83 & 12.00 \\
Ground truth & 20.88 & 41.33 & 16.40 \\
Sibling solution & 46.38 & 22.67 & 19.60 \\
Hint & \textbf{62.50} & \textbf{44.00} & \textbf{24.20} \\
\bottomrule
\end{tabular}
\caption{Average validation accuracy (\%) across grid sizes under different privileged contexts for failed trajectories. The successful-trajectory branch remains unchanged across all conditions. Boldface indicates the best result in each column.}
\label{tab:privileged_context}
\end{table}

\subsection{When Does Rephrasing Help?}
\label{sec:rephrasing}

\begin{table}[!t]
\centering
\setlength{\tabcolsep}{4pt}
\begin{tabular*}{\columnwidth}{@{\extracolsep{\fill}}lcc@{}}
\toprule
Update Strategy
& \multicolumn{1}{c}{Sudoku} 
& \multicolumn{1}{c}{Calcudoku} \\
& \multicolumn{1}{c}{(NR / R)} 
& \multicolumn{1}{c}{(NR / R)} \\
\midrule
Magnitude Only
& 27.50 / \textbf{55.00}
& 41.83 / \textbf{43.17} \\
Reverse-KL Only
& \textbf{47.75} / 42.50
& \textbf{42.50} / 28.67 \\
H$^2$SD
& 61.50 / \textbf{62.50}
& 43.17 / \textbf{44.00} \\
\bottomrule
\end{tabular*}
\caption{Effect of successful-trajectory rephrasing on accuracy (\%). NR and R denote training without and with rephrasing, respectively. Boldface indicates the better result within each pair.}
\label{tab:rephrasing}
\end{table}

Table~\ref{tab:rephrasing} holds the verified successful student response in the teacher context fixed and adds or removes only the rephrasing instruction, while the failure-side context is unchanged. The same rephrasing instruction improves Magnitude Only by 27.50 and 1.34 percentage points on Sudoku and Calcudoku, and H$^2$SD by 1.00 and 0.83 points, but reduces Reverse-KL Only by 5.25 and 13.83 points.

Many reasoning problems such as Calcudoku admit multiple valid solutions, so an externally generated reference rationale may follow a reasoning path different from the correct trajectory taken by the student. Conditioning the teacher on the verified student response avoids this mismatch, while the rephrasing instruction mainly changes how the original trajectory is evaluated rather than replacing it with an alternative solution path. As illustrated in Figure~\ref{fig:teacher_student_schematic}, teachers in the rephrase group are more likely to reinforce correct decisions at key reasoning branches, such as number placements in Sudoku, while suppressing redundant or repetitive checking expressions such as ``but'' and ``wait''. This helps explain why rephrasing achieves higher accuracy with shorter responses. Importantly, however, the effect of this teacher signal depends on how it is incorporated into training. Under magnitude modulation, the teacher signal primarily emphasizes tokens that contribute to the correct solution without substantially altering the original policy distribution, thereby preserving the student's valid reasoning trajectory. In contrast, direct reverse-KL matching may overconstrain an already correct trajectory by forcing it toward the rephrase-conditioned teacher distribution. Figure~\ref{fig:routing_entropy} supports this interpretation: adding rephrasing to Reverse-KL Only causes actor entropy to decrease sharply, indicating a substantial loss of exploration, whereas Magnitude Only maintains relatively stable entropy.

\section{Conclusion}
We introduced H$^2$SD, a self-distillation framework that adapts teacher context and update strategy to trajectory correctness. It achieves strong performance across logic puzzles and mathematical reasoning tasks. Our analysis reveals that hint access alone does not explain these gains, reversed routing causes performance degradation and entropy collapse, and rephrasing benefits magnitude modulation but can reduce exploration under direct distribution matching. Future work will investigate extending H$^2$SD to agentic tasks with longer interaction trajectories and richer environmental feedback.

\section*{Limitations}
Our evaluation is limited to logic puzzles and mathematical reasoning tasks with reliable verifier-based correctness signals. Although the responses can be long in token length, they remain single-turn reasoning trajectories without repeated interaction with an external environment. It therefore remains unclear whether H$^2$SD generalizes to agentic settings that involve long-horizon decision-making, tool use, delayed feedback, or noisy environmental signals. In addition, our experiments focus on two Qwen3 models and use verifier-validated reference hints during training. Future work will study whether the same supervision strategy remains effective across broader model families and settings where privileged feedback is less complete or less reliable.

\section*{Broader Impacts and Responsible Research}
\label{sec:broader_impacts}

In this section, we discuss the broader implications, ethical considerations, and reproducibility of our work.

\paragraph{Ethical Considerations.}
All datasets utilized in this study are obtained from publicly available sources or procedurally generated and are free for academic research. Our data synthesis and evaluation processes do not involve the collection or processing of any sensitive personally identifiable information (PII). Consequently, we do not foresee any significant privacy risks associated with the data used in this work.

\paragraph{Reproducibility.}
We are committed to open-source and reproducible research.
We plan to release our training code, scripts for generating
training and validation data, and reference-hint generation
code. Detailed experimental configurations, hyperparameters,
and hardware specifications are provided in Appendix~\ref{sec:implementation-details},
while representative prompts and reference hints are
included in Appendix~\ref{app:prompts}.

\clearpage
\bibliographystyle{plainnat}
\bibliography{refs}

\begin{thebibliography}{22}
\providecommand{\natexlab}[1]{#1}
\providecommand{\url}[1]{\texttt{#1}}
\expandafter\ifx\csname urlstyle\endcsname\relax
  \providecommand{\doi}[1]{doi: #1}\else
  \providecommand{\doi}{doi: \begingroup \urlstyle{rm}\Url}\fi

\bibitem[Agarwal et~al.(2024)Agarwal, Vieillard, Zhou, Stanczyk, Ramos~Garea,
  Geist, and Bachem]{opdllm}
Rishabh Agarwal, Nino Vieillard, Yongchao Zhou, Piotr Stanczyk, Sabela
  Ramos~Garea, Matthieu Geist, and Olivier Bachem.
\newblock On-policy distillation of language models: Learning from
  self-generated mistakes.
\newblock In B.~Kim, Y.~Yue, S.~Chaudhuri, K.~Fragkiadaki, M.~Khan, and Y.~Sun,
  editors, \emph{International Conference on Learning Representations}, volume
  2024, pages 21246--21263, 2024.
\newblock URL
  \url{https://proceedings.iclr.cc/paper_files/paper/2024/file/5be69a584901a26c521c2b51e40a4c20-Paper-Conference.pdf}.

\bibitem[{GLM-5 Team} et~al.(2026){GLM-5 Team}, Zeng, Lv, Hou, Du, Zheng, Chen,
  Yin, Ge, Huang, Xie, Zhu, Yin, Wang, Pan, Zeng, Zhang, Wang, Chen, Zhang,
  Jiao, Guo, Wang, Du, Wu, Wang, Li, Fan, Zhong, Liu, Zhao, Du, Dong, Lu,
  Shuang-Li, Cao, Liu, Jiang, Chen, Zhang, Huang, Dong, Xu, Wei, An, Niu, Zhu,
  Wen, Cen, Bai, Qiao, Wang, Wang, Zhu, Liu, Li, Wang, Wen, Huang, Cai, Yu, Li,
  Hu, Zhang, Zhang, Lin, Yang, Wang, Ai, Zhu, Yi, Chen, Wen, Sun, Zhao, Hu,
  Zhang, Liu, Zhang, Peng, Tai, Zhang, Liu, Wang, Yan, Ge, Liu, Chu, Zhao,
  Wang, Zhao, Ren, Wang, Zhang, Gui, Zhao, Li, An, Li, Yuan, Du, Liu, Zhi,
  Duan, Zhou, Wei, Wang, Luo, Zhang, Sha, Xu, Wu, Ding, Chen, Li, Lin, Ta, Zou,
  Song, Yang, Tu, Yang, Wu, Zhang, Li, Li, Fan, Qin, Tian, Zhang, Yu, Liang,
  Kuang, Cheng, Li, Yan, Hu, Ling, Fan, Xia, Zhang, Zhang, Pan, Zou, Zhang,
  Liu, Wu, Li, Wang, Zhu, Tan, Zhou, Pan, Zhang, Su, Geng, Yan, Tan, Bi, Shen,
  Yang, Li, Liu, Wang, Li, Wu, Zhang, Duan, Zhang, Liu, Jiang, Yan, Zhang, Wei,
  Chen, Feng, Yao, Chai, Wang, Zhang, Xu, Huang, Wang, Li, Dong, and
  Tang]{glm5}
{GLM-5 Team}, Aohan Zeng, Xin Lv, Zhenyu Hou, Zhengxiao Du, Qinkai Zheng, Bin
  Chen, Da~Yin, Chendi Ge, Chenghua Huang, Chengxing Xie, Chenzheng Zhu,
  Congfeng Yin, Cunxiang Wang, Gengzheng Pan, Hao Zeng, Haoke Zhang, Haoran
  Wang, Huilong Chen, Jiajie Zhang, Jian Jiao, Jiaqi Guo, Jingsen Wang,
  Jingzhao Du, Jinzhu Wu, Kedong Wang, Lei Li, Lin Fan, Lucen Zhong, Mingdao
  Liu, Mingming Zhao, Pengfan Du, Qian Dong, Rui Lu, Shuang-Li, Shulin Cao,
  Song Liu, Ting Jiang, Xiaodong Chen, Xiaohan Zhang, Xuancheng Huang, Xuezhen
  Dong, Yabo Xu, Yao Wei, Yifan An, Yilin Niu, Yitong Zhu, Yuanhao Wen, Yukuo
  Cen, Yushi Bai, Zhongpei Qiao, Zihan Wang, Zikang Wang, Zilin Zhu, Ziqiang
  Liu, Zixuan Li, Bojie Wang, Bosi Wen, Can Huang, Changpeng Cai, Chao Yu, Chen
  Li, Chengwei Hu, Chenhui Zhang, Dan Zhang, Daoyan Lin, Dayong Yang, Di~Wang,
  Ding Ai, Erle Zhu, Fangzhou Yi, Feiyu Chen, Guohong Wen, Hailong Sun, Haisha
  Zhao, Haiyi Hu, Hanchen Zhang, Hanrui Liu, Hanyu Zhang, Hao Peng, Hao Tai,
  Haobo Zhang, He~Liu, Hongwei Wang, Hongxi Yan, Hongyu Ge, Huan Liu, Huanpeng
  Chu, Jia'ni Zhao, Jiachen Wang, Jiajing Zhao, Jiamin Ren, Jiapeng Wang,
  Jiaxin Zhang, Jiayi Gui, Jiayue Zhao, Jijie Li, Jing An, Jing Li, Jingwei
  Yuan, Jinhua Du, Jinxin Liu, Junkai Zhi, Junwen Duan, Kaiyue Zhou, Kangjian
  Wei, Ke~Wang, Keyun Luo, Laiqiang Zhang, Leigang Sha, Liang Xu, Lindong Wu,
  Lintao Ding, Lu~Chen, Minghao Li, Nianyi Lin, Pan Ta, Qiang Zou, Rongjun
  Song, Ruiqi Yang, Shangqing Tu, Shangtong Yang, Shaoxiang Wu, Shengyan Zhang,
  Shijie Li, Shuang Li, Shuyi Fan, Wei Qin, Wei Tian, Weining Zhang, Wenbo Yu,
  Wenjie Liang, Xiang Kuang, Xiangmeng Cheng, Xiangyang Li, Xiaoquan Yan,
  Xiaowei Hu, Xiaoying Ling, Xing Fan, Xingye Xia, Xinyuan Zhang, Xinze Zhang,
  Xirui Pan, Xu~Zou, Xunkai Zhang, Yadi Liu, Yandong Wu, Yanfu Li, Yidong Wang,
  Yifan Zhu, Yijun Tan, Yilin Zhou, Yiming Pan, Ying Zhang, Yinpei Su, Yipeng
  Geng, Yong Yan, Yonglin Tan, Yuean Bi, Yuhan Shen, Yuhao Yang, Yujiang Li,
  Yunan Liu, Yunqing Wang, Yuntao Li, Yurong Wu, Yutao Zhang, Yuxi Duan, Yuxuan
  Zhang, Zezhen Liu, Zhengtao Jiang, Zhenhe Yan, Zheyu Zhang, Zhixiang Wei,
  Zhuo Chen, Zhuoer Feng, Zijun Yao, Ziwei Chai, Ziyuan Wang, Zuzhou Zhang, Bin
  Xu, Minlie Huang, Hongning Wang, Juanzi Li, Yuxiao Dong, and Jie Tang.
\newblock {GLM-5}: from vibe coding to agentic engineering, 2026.
\newblock URL \url{https://arxiv.org/abs/2602.15763}.

\bibitem[Gu et~al.(2024)Gu, Dong, Wei, and Huang]{minillm}
Yuxian Gu, Li~Dong, Furu Wei, and Minlie Huang.
\newblock {MiniLLM}: Knowledge distillation of large language models.
\newblock In B.~Kim, Y.~Yue, S.~Chaudhuri, K.~Fragkiadaki, M.~Khan, and Y.~Sun,
  editors, \emph{International Conference on Learning Representations}, volume
  2024, pages 32694--32717, 2024.
\newblock URL
  \url{https://proceedings.iclr.cc/paper_files/paper/2024/file/8ac015d409635f196f9e3e9dcfb9a94e-Paper-Conference.pdf}.

\bibitem[Guo et~al.(2025)Guo, Yang, Zhang, Song, Wang, Zhu, Xu, Zhang, Ma, Bi,
  Zhang, Yu, Wu, Wu, Gou, Shao, Li, Gao, Liu, Xue, Wang, Wu, Feng, Lu, Zhao,
  Deng, Ruan, Dai, Chen, Ji, Li, Lin, Dai, Luo, Hao, Chen, Li, Zhang, Xu, Ding,
  Gao, Qu, Li, Guo, Li, Chen, Yuan, Tu, Qiu, Li, Cai, Ni, Liang, Chen, Dong,
  Hu, You, Gao, Guan, Huang, Yu, Wang, Zhang, Zhao, Wang, Zhang, Xu, Xia,
  Zhang, Zhang, Tang, Zhou, Li, Wang, Li, Tian, Huang, Zhang, Wang, Chen, Du,
  Ge, Zhang, Pan, Wang, Chen, Jin, Chen, Lu, Zhou, Chen, Ye, Wang, Yu, Zhou,
  Pan, Li, Zhou, Wu, Yun, Pei, Sun, Wang, Zeng, Liu, Liang, Gao, Yu, Zhang,
  Xiao, An, Liu, Wang, Chen, Nie, Cheng, Liu, Xie, Liu, Yang, Li, Su, Lin, Li,
  Jin, Shen, Chen, Sun, Wang, Song, Zhou, Wang, Shan, Li, Wang, Wei, Zhang, Xu,
  Li, Zhao, Sun, Wang, Yu, Zhang, Shi, Xiong, He, Piao, Wang, Tan, Ma, Liu,
  Guo, Ou, Wang, Gong, Zou, He, Xiong, Luo, You, Liu, Zhou, Zhu, Huang, Li,
  Zheng, Zhu, Ma, Tang, Zha, Yan, Ren, Ren, Sha, Fu, Xu, Xie, Zhang, Hao, Ma,
  Yan, Wu, Gu, Zhu, Liu, Li, Xie, Song, Pan, Huang, Xu, Zhang, and Zhang]{dsr1}
Daya Guo, Dejian Yang, Haowei Zhang, Junxiao Song, Peiyi Wang, Qihao Zhu,
  Runxin Xu, Ruoyu Zhang, Shirong Ma, Xiao Bi, Xiaokang Zhang, Xingkai Yu,
  Yu~Wu, Z.~F. Wu, Zhibin Gou, Zhihong Shao, Zhuoshu Li, Ziyi Gao, Aixin Liu,
  Bing Xue, Bingxuan Wang, Bochao Wu, Bei Feng, Chengda Lu, Chenggang Zhao,
  Chengqi Deng, Chong Ruan, Damai Dai, Deli Chen, Dongjie Ji, Erhang Li,
  Fangyun Lin, Fucong Dai, Fuli Luo, Guangbo Hao, Guanting Chen, Guowei Li,
  H.~Zhang, Hanwei Xu, Honghui Ding, Huazuo Gao, Hui Qu, Hui Li, Jianzhong Guo,
  Jiashi Li, Jingchang Chen, Jingyang Yuan, Jinhao Tu, Junjie Qiu, Junlong Li,
  J.~L. Cai, Jiaqi Ni, Jian Liang, Jin Chen, Kai Dong, Kai Hu, Kaichao You,
  Kaige Gao, Kang Guan, Kexin Huang, Kuai Yu, Lean Wang, Lecong Zhang, Liang
  Zhao, Litong Wang, Liyue Zhang, Lei Xu, Leyi Xia, Mingchuan Zhang, Minghua
  Zhang, Minghui Tang, Mingxu Zhou, Meng Li, Miaojun Wang, Mingming Li, Ning
  Tian, Panpan Huang, Peng Zhang, Qiancheng Wang, Qinyu Chen, Qiushi Du, Ruiqi
  Ge, Ruisong Zhang, Ruizhe Pan, Runji Wang, R.~J. Chen, R.~L. Jin, Ruyi Chen,
  Shanghao Lu, Shangyan Zhou, Shanhuang Chen, Shengfeng Ye, Shiyu Wang,
  Shuiping Yu, Shunfeng Zhou, Shuting Pan, S.~S. Li, Shuang Zhou, Shaoqing Wu,
  Tao Yun, Tian Pei, Tianyu Sun, T.~Wang, Wangding Zeng, Wen Liu, Wenfeng
  Liang, Wenjun Gao, Wenqin Yu, Wentao Zhang, W.~L. Xiao, Wei An, Xiaodong Liu,
  Xiaohan Wang, Xiaokang Chen, Xiaotao Nie, Xin Cheng, Xin Liu, Xin Xie,
  Xingchao Liu, Xinyu Yang, Xinyuan Li, Xuecheng Su, Xuheng Lin, X.~Q. Li,
  Xiangyue Jin, Xiaojin Shen, Xiaosha Chen, Xiaowen Sun, Xiaoxiang Wang, Xinnan
  Song, Xinyi Zhou, Xianzu Wang, Xinxia Shan, Y.~K. Li, Y.~Q. Wang, Y.~X. Wei,
  Yang Zhang, Yanhong Xu, Yao Li, Yao Zhao, Yaofeng Sun, Yaohui Wang, Yi~Yu,
  Yichao Zhang, Yifan Shi, Yiliang Xiong, Ying He, Yishi Piao, Yisong Wang,
  Yixuan Tan, Yiyang Ma, Yiyuan Liu, Yongqiang Guo, Yuan Ou, Yuduan Wang, Yue
  Gong, Yuheng Zou, Yujia He, Yunfan Xiong, Yuxiang Luo, Yuxiang You, Yuxuan
  Liu, Yuyang Zhou, Y.~X. Zhu, Yanping Huang, Yaohui Li, Yi~Zheng, Yuchen Zhu,
  Yunxian Ma, Ying Tang, Yukun Zha, Yuting Yan, Z.~Z. Ren, Zehui Ren, Zhangli
  Sha, Zhe Fu, Zhean Xu, Zhenda Xie, Zhengyan Zhang, Zhewen Hao, Zhicheng Ma,
  Zhigang Yan, Zhiyu Wu, Zihui Gu, Zijia Zhu, Zijun Liu, Zilin Li, Ziwei Xie,
  Ziyang Song, Zizheng Pan, Zhen Huang, Zhipeng Xu, Zhongyu Zhang, and Zhen
  Zhang.
\newblock {DeepSeek-R1} incentivizes reasoning in {LLMs} through reinforcement
  learning.
\newblock \emph{Nature}, 645\penalty0 (8081):\penalty0 633--638, September
  2025.
\newblock ISSN 1476-4687.
\newblock \doi{10.1038/s41586-025-09422-z}.
\newblock URL \url{https://doi.org/10.1038/s41586-025-09422-z}.

\bibitem[He et~al.(2026)He, Kaur, Bhaskar, Yang, Liu, Ri, Fowl, Panigrahi,
  Chen, and Arora]{zero}
Yinghui He, Simran Kaur, Adithya Bhaskar, Yongjin Yang, Jiarui Liu, Narutatsu
  Ri, Liam Fowl, Abhishek Panigrahi, Danqi Chen, and Sanjeev Arora.
\newblock Self-distillation zero: Self-revision turns binary rewards into dense
  supervision, 2026.
\newblock URL \url{https://arxiv.org/abs/2604.12002}.

\bibitem[Heakl et~al.(2026)Heakl, Shaker, Mohamed, Elbadry, Fetouh, Khan, and
  Khan]{cepo}
Ahmed Heakl, Abdelrahman~M. Shaker, Youssef Mohamed, Rania Elbadry, Omar
  Fetouh, Fahad~Shahbaz Khan, and Salman Khan.
\newblock {CEPO}: {RLVR} self-distillation using contrastive evidence policy
  optimization, 2026.
\newblock URL \url{https://arxiv.org/abs/2605.19436}.

\bibitem[Huang et~al.(2026)Huang, Cheng, Zhao, Shen, Yang, Feng, Gu, Yu, and
  Qin]{golf}
Lei Huang, Xiang Cheng, Chenxiao Zhao, Guobin Shen, Junjie Yang, Xiaocheng
  Feng, Yuxuan Gu, Xing Yu, and Bing Qin.
\newblock Bootstrapping exploration with group-level natural language feedback
  in reinforcement learning, 2026.
\newblock URL \url{https://arxiv.org/abs/2603.04597}.

\bibitem[H\"{u}botter et~al.(2026)H\"{u}botter, L\"{u}beck, Behric, Baumann,
  Bagatella, Marta, Hakimi, Shenfeld, Kleine~Buening, Guestrin, and
  Krause]{sdpo}
Jonas H\"{u}botter, Frederike L\"{u}beck, Lejs~Deen Behric, Anton Baumann,
  Marco Bagatella, Daniel Marta, Ido Hakimi, Idan Shenfeld, Thomas
  Kleine~Buening, Carlos Guestrin, and Andreas Krause.
\newblock Reinforcement learning via self-distillation.
\newblock In Tong Zhang, Miroslav Dudik, Martin Jaggi, Alekh Agarwal, Sharon
  Li, Dale Schuurmans, Jerry Zhu, Felix Berkenkamp, Hanze Dong, and Alberto
  Bietti, editors, \emph{Proceedings of the 43rd International Conference on
  Machine Learning}, volume 306 of \emph{Proceedings of Machine Learning
  Research}, pages 48677--48721. PMLR, 06--11 Jul 2026.
\newblock URL \url{https://proceedings.mlr.press/v306/hubotter26a.html}.

\bibitem[Ke et~al.(2026)Ke, Wen, Li, He, and Zhang]{egrsd}
Junlong Ke, Zichen Wen, Weijia Li, Conghui He, and Linfeng Zhang.
\newblock Respecting self-uncertainty in on-policy self-distillation for
  efficient {LLM} reasoning, 2026.
\newblock URL \url{https://arxiv.org/abs/2605.13255}.

\bibitem[{Kimi Team} et~al.(2026){Kimi Team}, Bai, Bai, Bao, Cai, Cao, Chai,
  Charles, Che, Chen, Chen, Chen, Chen, Chen, Chen, Chen, Chen, Chen, Chen,
  Chen, Chen, Chen, Chen, Chen, Chen, Chen, Chen, Chen, Chen, Cheng, Cheng,
  Chu, Cui, Deng, Diao, Ding, Dong, Dong, Dong, Dong, Du, Du, Du, Du, Du, Fan,
  Fang, Feng, Feng, Fu, Fu, Gao, Gao, Ge, Geng, Gong, Gong, Gongque, Gu, Gu,
  Gu, Guan, Guan, Guo, Hao, He, He, He, He, He, He, Hong, Hu, Hu, Hu, Hu,
  Huang, Huang, Huang, Huang, Jia, Jiang, Jiang, Jin, Jing, Lai, Li, Li, Li,
  Li, Li, Li, Li, Li, Li, Li, Li, Li, Li, Li, Li, Li, Li, Li, Li, Li, Li, Li,
  Li, Li, Liao, Lin, Lin, Lin, Lin, Lin, Liu, Liu, Liu, Liu, Liu, Liu, Liu,
  Liu, Liu, Liu, Liu, Liu, Liu, Liu, Liu, Liu, Liu, Lu, Lu, Lu, Luo, Luo, Luo,
  Luo, Ma, Mao, Mei, Men, Meng, Meng, Miao, Ni, Ouyang, Pan, Pang, Qian, Qin,
  Qin, Qiu, Qu, Shang, Shao, Shen, Shen, Shi, Shi, Shi, Song, Song, Song, Song,
  Su, Su, Su, Sui, Sun, Sun, Sun, Sung, Tai, Tang, Tang, Tang, Tang, Tao, Teng,
  Tian, Tian, Wang, Wang, Wang, Wang, Wang, Wang, Wang, Wang, Wang, Wang, Wang,
  Wang, Wang, Wang, Wang, Wang, Wang, Wang, Wang, Wang, Wang, Wang, Wang, Wang,
  Wang, Wang, Wang, Wang, Wang, Wang, Wang, Wang, Wang, Wang, Wang, Wang, Wang,
  Wang, Wang, Wei, Wei, Wen, Wen, Wu, Wu, Wu, Wu, Wu, Wu, Wu, Wu, Wu, Xiao,
  Xie, Xie, Xie, Xing, Xu, Xu, Xu, Xu, Xu, Xu, Xu, Xu, Xu, Xu, Xu, Xu, Xu, Xu,
  Xu, Yan, Yan, Yang, Yang, Yang, Yang, Yang, Yang, Yang, Yang, Yang, Yang,
  Yang, Yang, Yang, Yang, Yao, Ye, Ye, Ye, Ye, Yebo, Yin, Yu, Yu, Yu, Yu, Yuan,
  Yuan, Yuan, Yue, Zeng, Zha, Zhan, Zhang, Zhang, Zhang, Zhang, Zhang, Zhang,
  Zhang, Zhang, Zhang, Zhang, Zhang, Zhang, Zhang, Zhang, Zhang, Zhang, Zhang,
  Zhang, Zhang, Zhao, Zhao, Zhao, Zhao, Zhao, Zhao, Zhao, Zhao, Zheng, Zheng,
  Zheng, Zheng, Zhong, Zhong, Zhong, Zhou, Zhou, Zhou, Zhou, Zhu, Zhu, Zhu,
  Zhu, Zhu, Zhuang, Zhuang, Zou, and Zu]{kimi2.5}
{Kimi Team}, Tongtong Bai, Yifan Bai, Yiping Bao, S.~H. Cai, Yuan Cao, Ziwei
  Chai, Y.~Charles, H.~S. Che, Cheng Chen, Guanduo Chen, Huarong Chen, Jia
  Chen, Jianlong Chen, Jun Chen, Kefan Chen, Liang Chen, Ruijue Chen, Xinhao
  Chen, Yanru Chen, Yanxu Chen, Yicun Chen, Yimin Chen, Yingjiang Chen, Yuankun
  Chen, Yujie Chen, Yutian Chen, Zhirong Chen, Ziwei Chen, Dazhi Cheng, Yean
  Cheng, Minghan Chu, Jialei Cui, Jiaqi Deng, Muxi Diao, Hao Ding, Mengfan
  Dong, Mengnan Dong, Yuxin Dong, Yuhao Dong, Angang Du, Chenzhuang Du, Dikang
  Du, Lingxiao Du, Yulun Du, Yu~Fan, Shengjun Fang, Qiulin Feng, Yichen Feng,
  Garimugai Fu, Kelin Fu, Hongcheng Gao, Tong Gao, Yuyao Ge, Shangyi Geng,
  Chengyang Gong, Xiaochen Gong, Zhuoma Gongque, Qizheng Gu, Xinran Gu, Yicheng
  Gu, Longyu Guan, Shuhao Guan, Yuanying Guo, Xiaoru Hao, Dailan He, Tianhong
  He, Weiran He, Wenyang He, Yibo He, Yunjia He, Chao Hong, Hao Hu, Jiaxi Hu,
  Yangyang Hu, Zhenxing Hu, Ke~Huang, Ruiyuan Huang, Weixiao Huang, Zhiqi
  Huang, Chaobo Jia, Tao Jiang, Zhejun Jiang, Xinyi Jin, Yu~Jing, Guokun Lai,
  Aidi Li, C.~Li, Cheng Li, Fang Li, Guanghe Li, Guanyu Li, Haitao Li, Haoyang
  Li, Jia Li, Jingwei Li, Junxiong Li, Lincan Li, Mo~Li, Weihong Li, Wentao Li,
  Xinhang Li, Xinhao Li, Yang Li, Yanhao Li, Yiwei Li, Yuxiao Li, Zhaowei Li,
  Zhaoxi Li, Zheming Li, Weilong Liao, Jiawei Lin, Xiaohan Lin, Yibo Lin,
  Zhishan Lin, Zichao Lin, Cheng Liu, Chenyu Liu, Hongzhang Liu, Liang Liu,
  Shaowei Liu, Shudong Liu, Shuran Liu, Tianwei Liu, Tianyu Liu, Weizhou Liu,
  Xiangyan Liu, Yangyang Liu, Yanming Liu, Yibo Liu, Yuanxin Liu, Zhengying
  Liu, Zhongnuo Liu, Enzhe Lu, Haoyu Lu, Zhiyuan Lu, G.~Luo, Junyu Luo, Tongxu
  Luo, Yashuo Luo, Long Ma, Shaoguang Mao, Yuan Mei, Xin Men, Fanqing Meng,
  Zhiyong Meng, Yibo Miao, Minqing Ni, Kun Ouyang, Siyuan Pan, Bo~Pang, Yuchao
  Qian, Ruoyu Qin, Zeyu Qin, Jiezhong Qiu, Bowen Qu, Zeyu Shang, Youbo Shao,
  Tianxiao Shen, Zhennan Shen, Juanfeng Shi, Lidong Shi, Shengyuan Shi, Feifan
  Song, Pengwei Song, Tianhui Song, Xiaoxi Song, Hongjin Su, Jianlin Su,
  Zhaochen Su, Lin Sui, Jinsong Sun, Junyao Sun, Tongyu Sun, Flood Sung,
  Yunpeng Tai, Chuning Tang, Heyi Tang, Xiaojuan Tang, Zhengyang Tang, Jiawen
  Tao, Shiyuan Teng, Chaoran Tian, Pengfei Tian, Bowen Wang, Chensi Wang,
  Chuang Wang, Congcong Wang, Dingkun Wang, Dinglu Wang, Dongliang Wang, Feng
  Wang, Hailong Wang, Haiming Wang, Hao Wang, Hengzhi Wang, Huaqing Wang, Hui
  Wang, Jiahao Wang, Jinhong Wang, Jiuzheng Wang, Kaixin Wang, Linian Wang,
  Qibin Wang, Shengjie Wang, Shuyi Wang, Si~Wang, Wei Wang, Xiaochen Wang,
  Xinyuan Wang, Yao Wang, Yejie Wang, Yipu Wang, Yiqin Wang, Yucheng Wang,
  Yuzhi Wang, Zhaoji Wang, Zhaowei Wang, Zhengtao Wang, Zhexu Wang, Zifan Wang,
  Zihan Wang, Zizhe Wang, Chu Wei, Ming Wei, Chuan Wen, Zichen Wen, Chengjie
  Wu, Haoning Wu, Junyan Wu, Rucong Wu, Wenhao Wu, Yuefeng Wu, Yuhao Wu, Yuxin
  Wu, Zijian Wu, Chenjun Xiao, Jin Xie, Xiaotong Xie, Yuchong Xie, Bowei Xing,
  Boyu Xu, Jianfan Xu, Jing Xu, Jinjing Xu, L.~H. Xu, Lin Xu, Suting Xu, Weixin
  Xu, Xinbo Xu, Xinran Xu, Yangchuan Xu, Yichang Xu, Yuemeng Xu, Zelai Xu,
  Ziyao Xu, Junjie Yan, Yuzi Yan, Guangyao Yang, Hao Yang, Junwei Yang, Kai
  Yang, Ningyuan Yang, Xiaofei Yang, Xinlong Yang, Xinyu Yang, Ying Yang,
  Yi~Yang, Yi~Yang, Zhen Yang, Zhilin Yang, Zonghan Yang, Haotian Yao, Dan Ye,
  Haoran Ye, Wenjie Ye, Zhuorui Ye, Peng Yebo, Bohong Yin, Chengzhen Yu,
  Longhui Yu, Tao Yu, Tianxiang Yu, Enming Yuan, Mengjie Yuan, Xiaokun Yuan,
  Yang Yue, Weihao Zeng, Dunyuan Zha, Haobing Zhan, Dehao Zhang, Hao Zhang, Jin
  Zhang, Puqi Zhang, Qiao Zhang, Rui Zhang, Xiaobin Zhang, Xiaoyun Zhang,
  Y.~Zhang, Yadong Zhang, Yangkun Zhang, Yichi Zhang, Yizhi Zhang, Yongting
  Zhang, Yu~Zhang, Yushun Zhang, Yutao Zhang, Yutong Zhang, Zheng Zhang,
  Chenguang Zhao, Feifan Zhao, Jinxiang Zhao, Shuai Zhao, Xiangyu Zhao, Xuanle
  Zhao, Yikai Zhao, Zijia Zhao, Huabin Zheng, Ruihan Zheng, Shaojie Zheng,
  Tengyang Zheng, Junfeng Zhong, Longguang Zhong, Weiming Zhong, M.~Zhou,
  Runjie Zhou, Xinyu Zhou, Zaida Zhou, Jinguo Zhu, Liya Zhu, Xinhao Zhu, Yuxuan
  Zhu, Zhen Zhu, Jingze Zhuang, Weiyu Zhuang, Ying Zou, and Xinxing Zu.
\newblock {Kimi K2.5}: Visual agentic intelligence, 2026.
\newblock URL \url{https://arxiv.org/abs/2602.02276}.

\bibitem[Li et~al.(2026)Li, Yang, Fang, Song, Zheng, Guo, Zhang, Wang, and
  Chua]{srpo}
Gengsheng Li, Tianyu Yang, Junfeng Fang, Mingyang Song, Mao Zheng, Haiyun Guo,
  Dan Zhang, Jinqiao Wang, and Tat-Seng Chua.
\newblock Unifying group-relative and self-distillation policy optimization via
  sample routing, 2026.
\newblock URL \url{https://arxiv.org/abs/2604.02288}.

\bibitem[Li et~al.(2025)Li, Zou, and Liu]{torl}
Xuefeng Li, Haoyang Zou, and Pengfei Liu.
\newblock {ToRL}: Scaling tool-integrated {RL}, 2025.
\newblock URL \url{https://arxiv.org/abs/2503.23383}.

\bibitem[Lu and {Thinking Machines Lab}(2025)]{opd-thinkingmachine}
Kevin Lu and {Thinking Machines Lab}.
\newblock On-policy distillation.
\newblock \emph{Thinking Machines Lab: Connectionism}, 2025.
\newblock \doi{10.64434/tml.20251026}.
\newblock URL \url{https://thinkingmachines.ai/blog/on-policy-distillation/}.

\bibitem[Qian et~al.(2025)Qian, Acikgoz, He, WANG, Chen, Hakkani-Tur, Tur, and
  Ji]{toolrl}
Cheng Qian, Emre~Can Acikgoz, Qi~He, Hongru WANG, Xiusi Chen, Dilek
  Hakkani-Tur, Gokhan Tur, and Heng Ji.
\newblock {ToolRL}: Reward is all tool learning needs.
\newblock In D.~Belgrave, C.~Zhang, H.~Lin, R.~Pascanu, P.~Koniusz,
  M.~Ghassemi, and N.~Chen, editors, \emph{Advances in Neural Information
  Processing Systems}, volume 38, Main Conference, pages 105523--105553. Curran
  Associates, Inc., 2025.
\newblock \doi{10.52202/085713-3524}.
\newblock URL
  \url{https://proceedings.neurips.cc/paper_files/paper/2025/file/97c5b2707228e7e3fb67e4ecc2e0e607-Paper-Conference.pdf}.

\bibitem[Shao et~al.(2024)Shao, Wang, Zhu, Xu, Song, Bi, Zhang, Zhang, Li, Wu,
  and Guo]{grpo}
Zhihong Shao, Peiyi Wang, Qihao Zhu, Runxin Xu, Junxiao Song, Xiao Bi, Haowei
  Zhang, Mingchuan Zhang, Y.~K. Li, Y.~Wu, and Daya Guo.
\newblock {DeepSeekMath}: Pushing the limits of mathematical reasoning in open
  language models, 2024.
\newblock URL \url{https://arxiv.org/abs/2402.03300}.

\bibitem[Shenfeld et~al.(2026)Shenfeld, Damani, H\"{u}botter, and
  Agrawal]{sdft}
Idan Shenfeld, Mehul Damani, Jonas H\"{u}botter, and Pulkit Agrawal.
\newblock Self-distillation enables continual learning.
\newblock In Tong Zhang, Miroslav Dudik, Martin Jaggi, Alekh Agarwal, Sharon
  Li, Dale Schuurmans, Jerry Zhu, Felix Berkenkamp, Hanze Dong, and Alberto
  Bietti, editors, \emph{Proceedings of the 43rd International Conference on
  Machine Learning}, volume 306 of \emph{Proceedings of Machine Learning
  Research}, pages 111542--111558. PMLR, 06--11 Jul 2026.
\newblock URL \url{https://proceedings.mlr.press/v306/shenfeld26a.html}.

\bibitem[Stein et~al.(2026)Stein, Huang, and Goldstein]{gates}
Alex Stein, Furong Huang, and Tom Goldstein.
\newblock {GATES}: Self-distillation under privileged context with consensus
  gating, 2026.
\newblock URL \url{https://arxiv.org/abs/2602.20574}.

\bibitem[Yang et~al.(2026)Yang, Qin, Si, Chen, Gu, Yao, Lin, Wang, Wang, and
  Duan]{rlsd}
Chenxu Yang, Chuanyu Qin, Qingyi Si, Minghui Chen, Naibin Gu, Dingyu Yao, Zheng
  Lin, Weiping Wang, Jiaqi Wang, and Nan Duan.
\newblock Self-distilled {RLVR}, 2026.
\newblock URL \url{https://arxiv.org/abs/2604.03128}.

\bibitem[Ye et~al.(2026)Ye, Dong, Wu, Huang, and Wei]{opcd}
Tianzhu Ye, Li~Dong, Xun Wu, Shaohan Huang, and Furu Wei.
\newblock On-policy context distillation for language models, 2026.
\newblock URL \url{https://arxiv.org/abs/2602.12275}.

\bibitem[Yu et~al.(2025)Yu, Zhang, Zhu, Yuan, Zuo, Yue, Dai, Fan, Liu, liu,
  Liu, Liu, Lin, Lin, Ma, Sheng, Tong, Zhang, Zhang, Zhang, Zhang, Zhu, Zhu,
  Chen, Chen, Wang, Yu, Song, Wei, Zhou, Liu, Ma, Zhang, Yan, Wu, and
  Wang]{dapo}
Qiying Yu, Zheng Zhang, Ruofei Zhu, Yufeng Yuan, Xiaochen Zuo, Yu~Yue, Weinan
  Dai, Tiantian Fan, Gaohong Liu, juncai liu, LingJun Liu, Xin Liu, Haibin Lin,
  Zhiqi Lin, Bole Ma, Guangming Sheng, Yuxuan Tong, Chi Zhang, Mofan Zhang,
  Ru~Zhang, Wang Zhang, Hang Zhu, Jinhua Zhu, Jiaze Chen, Jiangjie Chen,
  Chengyi Wang, Hongli Yu, Yuxuan Song, Xiangpeng Wei, Hao Zhou, Jingjing Liu,
  Wei-Ying Ma, Ya-Qin Zhang, Lin Yan, Yonghui Wu, and Mingxuan Wang.
\newblock {DAPO}: An open-source {LLM} reinforcement learning system at scale.
\newblock In D.~Belgrave, C.~Zhang, H.~Lin, R.~Pascanu, P.~Koniusz,
  M.~Ghassemi, and N.~Chen, editors, \emph{Advances in Neural Information
  Processing Systems}, volume 38, Main Conference, pages 113222--113244. Curran
  Associates, Inc., 2025.
\newblock \doi{10.52202/085713-3775}.
\newblock URL
  \url{https://proceedings.neurips.cc/paper_files/paper/2025/file/a4277440d50f1f15d2cb4c14f7e0c0d2-Paper-Conference.pdf}.

\bibitem[Zhao et~al.(2026)Zhao, Xie, Liu, Huang, Pang, Chen, and Grover]{opsd}
Siyan Zhao, Zhihui Xie, Mengchen Liu, Jing Huang, Guan Pang, Feiyu Chen, and
  Aditya Grover.
\newblock Self-distilled reasoner: On-policy self-distillation for large
  language models.
\newblock In Tong Zhang, Miroslav Dudik, Martin Jaggi, Alekh Agarwal, Sharon
  Li, Dale Schuurmans, Jerry Zhu, Felix Berkenkamp, Hanze Dong, and Alberto
  Bietti, editors, \emph{Proceedings of the 43rd International Conference on
  Machine Learning}, volume 306 of \emph{Proceedings of Machine Learning
  Research}, pages 162433--162448. PMLR, 06--11 Jul 2026.
\newblock URL \url{https://proceedings.mlr.press/v306/zhao26be.html}.

\bibitem[Zheng et~al.(2025)Zheng, Liu, Li, Chen, Yu, Gao, Dang, Liu, Men, Yang,
  Zhou, and Lin]{gspo}
Chujie Zheng, Shixuan Liu, Mingze Li, Xiong-Hui Chen, Bowen Yu, Chang Gao, Kai
  Dang, Yuqiong Liu, Rui Men, An~Yang, Jingren Zhou, and Junyang Lin.
\newblock Group sequence policy optimization, 2025.
\newblock URL \url{https://arxiv.org/abs/2507.18071}.

\end{thebibliography}


\clearpage
\appendix
\normalsize
\clearpage
\normalsize
\raggedbottom

\section{Implementation and Evaluation Details}
\label{sec:implementation-details}
This appendix documents the experimental setup and evaluation, followed by the privileged contexts and prompts (Appendix~\ref{app:prompts}), local update analysis (Appendix~\ref{app:local_derivations}), token-level rephrasing analysis (Appendix~\ref{app:rephrase_token_scores}), and additional training diagnostics (Appendix~\ref{app:training_diagnostics}).

\subsection{Models, Data, and Training Setup}
The reported experiments use Qwen3-30B-A3B-Instruct-2507 (denoted Qwen3-30B-A3B-Instruct) and Qwen3-14B. The logic-task hyperparameters in Tables~\ref{tab:grpo-hyperparameters}--\ref{tab:opsd-hyperparameters} are shared by both models. The logic tasks use procedurally generated training
and validation sets. Each task family is trained independently, with different problem sizes and difficulty levels mixed within the same training set. Each family contains 1,500 training problems, while
the validation sets contain 800 Sudoku, 600 Calcudoku, and 1000 Arrow Maze problems, respectively. Both training and validation sets are evenly distributed across problem sizes and difficulty levels within each family. Math training uses DAPO-Math-17k (Yu et al., 2025); evaluation uses AIME 25, AIME 26, and Minerva Math.

All experiments use VERL version 0.7.0.dev0 and SGLang version 0.5.2. A typical task-level training run requires approximately 48 GPU-hours, although the computational cost varies across methods and tasks due to differences in training duration. H$^2$SD uses a frozen copy $\theta_0$ of the initial policy as its self-teacher. This copy is neither updated nor synchronized with the student. Kimi-K2.6 supplies offline reference hints; it is not the model that produces token-level teacher distributions during optimization.

Training horizons differ across methods because runs were stopped once validation performance had plateaued and further training showed no meaningful improvement. We do not enforce an identical number of optimization steps across methods.

\subsection{Evaluation and Aggregation}
\label{app:eval_protocol}

For each training run, we select the checkpoint with
the highest validation accuracy under greedy decoding.
For evaluation, we generate $S$ responses $y_{i,s}$
for each problem $x_i$ in an evaluation set of size
$N$, where $S=1$ for logic puzzles and $S=16$ for
math benchmarks. Let $V(x_i,y_{i,s})\in\{0,1\}$
denote acceptance by the task verifier. The reported
accuracy is
\begin{equation}
\widehat{\mathrm{Acc}}_S
=\frac{100}{NS}\sum_{i=1}^{N}\sum_{s=1}^{S}
V(x_i,y_{i,s}).
\label{eq:eval_accuracy}
\end{equation}
This measures the mean verifier acceptance rate
over problems and generated responses.

Family averages give equal weight to the two Sudoku sizes, two Calcudoku sizes, five Arrow Maze sizes, or three math benchmarks. Table~\ref{tab:main_results}'s Overall Avg. then gives equal weight to the four task families. The summary across model scales averages the two model-level results. Printed scores are rounded to two decimal places. All experiments use random seed 42.

\subsection{Baseline Definitions and Hyperparameters}
\label{sec:method-specific-hyperparameters}
We implement all baselines following their original papers unless otherwise specified. Any adaptations made for controlled comparison, including the use of shared offline reference hints, are explicitly documented below.

RLSD and SRPO throughout the results denote their hint-conditioned adaptations. RLSD replaces final-answer context with an offline reference hint and uses $\lambda=1$. SRPO applies a clipped policy-gradient update to successful responses and hint-conditioned generalized Jensen--Shannon divergence (GJSD, $\alpha=0.5$) to failed responses. Offline hints do not require a correct sibling. OPSD uses hint-conditioned forward KL. H$^2$SD uses the verified response and a rephrasing instruction for successful-trajectory magnitude modulation, and an offline hint for failed-trajectory reverse KL.

The tables report the implemented logic-task settings rather than the defaults of the source papers. Recipes differ in clipping and divergence; OPSD uses $G=1$, versus $G=8$ for the other methods. Equal optimization steps therefore do not imply equal sampled-token or compute budgets. The reported runs also have different optimization horizons because training was continued only while validation performance was still improving. IS denotes importance sampling.

\paragraph{Optimizer settings.}
All methods use AdamW with a weight decay of $0.01$.
The learning rates of the baseline methods follow their
original papers. Table~\ref{tab:optimizer-settings}
summarizes the optimizer configurations for the logic tasks.

\begin{table}[!htbp]
\centering
\small
\setlength{\tabcolsep}{5pt}
\begin{tabular}{lcc}
\toprule
Method & Learning rate & Warmup steps \\
\midrule
GRPO & $1\times10^{-6}$ & 0 \\
OPSD & $5\times10^{-6}$ & 0 \\
RLSD+Hint & $1\times10^{-6}$ & 0 \\
SRPO+Hint & $5\times10^{-6}$ & 10 \\
H$^2$SD & $5\times10^{-6}$ & 0 \\
\bottomrule
\end{tabular}
\caption{Optimizer configurations for the logic tasks.
All methods use AdamW with a weight decay of $0.01$.
SRPO+Hint uses a 10-step linear warmup followed by
a constant learning rate, while the remaining methods
use constant learning rates without warmup.}
\label{tab:optimizer-settings}
\end{table}

\paragraph{Vocabulary approximation.}
The documented top-$K$ setting is 100 for OPSD, SRPO, and the H$^2$SD failure branch. For OPSD, the vocabulary partition retains the student's top-100 tokens and a residual-mass bucket.

\paragraph{Offline hint settings.}
Table~\ref{tab:hint-generation-settings} gives the generation parameters. Its temperature and token limit apply to hint construction, not to evaluation decoding. Appendix~\ref{sec:offline-hint-generation} describes the verification and use of these hints.

\begin{table}[!htbp]
\centering
\small
\setlength{\tabcolsep}{4pt}
\begin{tabularx}{\linewidth}
{@{}p{0.55\linewidth}>{\centering\arraybackslash}X@{}}
\toprule
\textbf{Parameter} & \textbf{GRPO} \\
\midrule
\multicolumn{2}{@{}l}{\textbf{Batching}} \\
Question batch size $B$ & 8 \\
Rollouts per prompt $G$ & 8 \\
\midrule
\multicolumn{2}{@{}l}{\textbf{Training and Loss}} \\
Gradient clip norm & 1.0 \\
Training objective & Clipped policy gradient \\
PPO clip & 0.20 \\
\bottomrule
\end{tabularx}
\caption{Logic hyperparameters for GRPO, shared by both models.}
\label{tab:grpo-hyperparameters}
\end{table}

\begin{table}[!htbp]
\centering
\small
\setlength{\tabcolsep}{4pt}
\begin{tabularx}{\linewidth}
{@{}p{0.55\linewidth}>{\centering\arraybackslash}X@{}}
\toprule
\textbf{Parameter} & \textbf{RLSD (with hint)} \\
\midrule
\multicolumn{2}{@{}l}{\textbf{Batching}} \\
Question batch size $B$ & 8 \\
Rollouts per prompt $G$ & 8 \\
\midrule
\multicolumn{2}{@{}l}{\textbf{Training}} \\
Gradient clip norm & 1.0 \\
\midrule
\multicolumn{2}{@{}l}{\textbf{RLSD (with hint) Loss}} \\
Teacher context & Offline Kimi-K2.6 hint \\
Training objective & RLSD magnitude modulation \\
Magnitude $\lambda$ / threshold $\epsilon_w$ & $1.0/0.2$ \\
\bottomrule
\end{tabularx}
\caption{Logic hyperparameters for RLSD (with hint), shared by both models.}
\label{tab:rlsd-hint-hyperparameters}
\end{table}

\begin{table}[!htbp]
\centering
\small
\setlength{\tabcolsep}{4pt}
\begin{tabularx}{\linewidth}
{@{}p{0.55\linewidth}>{\centering\arraybackslash}X@{}}
\toprule
\textbf{Parameter} & \textbf{H$^2$SD} \\
\midrule
\multicolumn{2}{@{}l}{\textbf{Batching}} \\
Question batch size $B$ & 8 \\
Rollouts per prompt $G$ & 8 \\
\midrule
\multicolumn{2}{@{}l}{\textbf{Training}} \\
Gradient clip norm & 1.0 \\
\midrule
\multicolumn{2}{@{}l}{\textbf{H$^2$SD Loss}} \\
Successful context & Verified response + rephrasing instruction \\
Successful objective & RLSD magnitude modulation \\
Failed context & Offline Kimi-K2.6 hint \\
Failed objective & Reverse KL \\
Magnitude $\lambda$ / threshold $\epsilon_w$ & $1.0/0.2$ \\
Reverse-KL coefficient $\gamma$ & 1.0 \\
Top-$K$ approximation for failed trajectories & 100 \\
\bottomrule
\end{tabularx}
\caption{Logic hyperparameters for H$^2$SD, shared by both models.}
\label{tab:h2sd-hyperparameters}
\end{table}

\begin{table}[!htbp]
\centering
\small
\setlength{\tabcolsep}{4pt}
\begin{tabularx}{\linewidth}
{@{}p{0.55\linewidth}>{\centering\arraybackslash}X@{}}
\toprule
\textbf{Parameter} & \textbf{SRPO (with hint)} \\
\midrule
\multicolumn{2}{@{}l}{\textbf{Batching}} \\
Question batch size $B$ & 8 \\
Rollouts per prompt $G$ & 8 \\
\midrule
\multicolumn{2}{@{}l}{\textbf{Training}} \\
Gradient clip norm & 1.0 \\
EMA teacher update rate & 0.05 \\
EMA update frequency & After each actor update \\
\midrule
\multicolumn{2}{@{}l}{\textbf{SRPO (with hint) Loss}} \\
Successful update & Clipped policy gradient \\
Failed context & Offline reference hint \\
Failed update & GJSD, $\alpha=0.5$ \\
Entropy weighting & $\exp(-\beta H_t)$ \\
Entropy coefficient $\beta$ & 1.0 \\
Entropy weight normalization & Unit mean over valid tokens \\
No-hint fallback & GRPO \\
Top-$K$ distillation & 100 \\
Rollout IS clip & 2.0 \\
\bottomrule
\end{tabularx}

\caption{Logic hyperparameters for hint-conditioned SRPO, shared by both models.}
\label{tab:srpo-hyperparameters}
\end{table}

\begin{table}[!htbp]
\centering
\small
\setlength{\tabcolsep}{4pt}
\begin{tabularx}{\linewidth}
{@{}p{0.55\linewidth}>{\centering\arraybackslash}X@{}}
\toprule
\textbf{Parameter} & \textbf{OPSD} \\
\midrule
\multicolumn{2}{@{}l}{\textbf{Batching}} \\
Question batch size $B$ & 8 \\
Rollouts per prompt $G$ & 1 \\
\midrule
\multicolumn{2}{@{}l}{\textbf{Training}} \\
Gradient clip norm & 0.1 \\
\midrule
\multicolumn{2}{@{}l}{\textbf{OPSD Loss}} \\
Teacher context & Offline Kimi-K2.6 hint \\
Distillation divergence & Forward KL \\
KL coefficient & 1.0 \\
Top-$K$ distillation & 100 \\
Pointwise KL clip & 0.05 \\
\bottomrule
\end{tabularx}
\caption{Logic hyperparameters for OPSD, shared by both models.}
\label{tab:opsd-hyperparameters}
\end{table}

\begin{table}[!htbp]
\centering
\small
\setlength{\tabcolsep}{4pt}
\begin{tabularx}{\linewidth}
{@{}p{0.55\linewidth}>{\centering\arraybackslash}X@{}}
\toprule
\textbf{Parameter} & \textbf{Hint Generation} \\
\midrule
\multicolumn{2}{@{}l}{\textbf{Generation}} \\
Generator & Kimi-K2.6 \\
Temperature & 0.1 \\
Maximum generation length & 65,536 tokens \\
Maximum attempts per problem & 5 \\
\bottomrule
\end{tabularx}
\caption{Offline reference-hint generation settings.}
\label{tab:hint-generation-settings}
\end{table}

\section{Privileged Contexts and Prompt Templates}
\label{app:prompts}
\subsection{Hint Generation and Successful-Response Scoring}
\label{sec:offline-hint-generation}
Kimi-K2.6 generates each reference hint offline, allowing up to five attempts. Only hints whose final answers pass the task verifier are retained. Hints condition the self-teacher during training and are unavailable to the student at inference.

Table~\ref{tab:prompt_templates} gives the prompts for a Sudoku-$6\times6$ task. In successful-response scoring, both teacher conditions receive the problem and the same correct student response. The rephrasing instruction changes teacher-forced scores of the original tokens; no rewritten response is generated. The 36-character answer format is specific to Sudoku-$6\times6$.

\begin{table*}[!t]
\centering
\begin{minipage}[t]{0.28\textwidth}
\vspace{0pt}
\begin{promptbox}[unbreakable,width=\linewidth,fontupper=\small,fonttitle=\small\bfseries]
{Hint Generation:\\System Prompt}

You are an expert Sudoku solver.

You will solve the given Sudoku puzzle from the beginning and summarize the whole standard reasoning process and provide the answer in the end, wrapped exactly in \texttt{<hint></hint>} tags:

\smallskip
\texttt{<hint>}

\texttt{...}

\texttt{</hint>}

\end{promptbox}

\begin{promptbox}[unbreakable,width=\linewidth,fontupper=\small,fonttitle=\small\bfseries]
{Hint Generation:\\User Prompt}

\texttt{\{prompt\}}

\end{promptbox}

\begin{promptbox}[unbreakable,width=\linewidth,fontupper=\small,fonttitle=\small\bfseries]
{Rephrasing:\\User Prompt}

\texttt{\{prompt\}}

\smallskip
Text to rephrase:

\smallskip
\texttt{\{first\_response\}}

\end{promptbox}

\end{minipage}
\hfill
\begin{minipage}[t]{0.68\textwidth}
\vspace{0pt}
\begin{promptbox}[unbreakable,width=\linewidth,fontupper=\small,fonttitle=\small\bfseries]
{Rephrasing: System Prompt}

You are a paraphraser for 6x6 Sudoku solution attempts.

Rewrite the input in different words while preserving the
same Sudoku-solving flow: checking fixed clues, applying
row, column, and box constraints, placing digits, and
reaching the final completed solution string.

Keep the useful reasoning steps that are necessary to
understand how the puzzle is solved, but you may simplify
or trim failed guesses, repeated checks, and minor
self-reflections that do not materially contribute to the
solution.

Preserve the original meaning, conclusions, puzzle facts,
digit placements, row/column/box references, and candidate
eliminations. Do not change any Sudoku clue, digit, or
coordinate.

\textbf{CRITICAL---the final answer block must be
byte-for-byte identical to the input:}

\begin{itemize}

    \item The input ends with exactly one
    \texttt{<answer>...</answer>} tag containing a
    36-character Sudoku solution string with no spaces,
    no newlines, and no separators between digits.

    \item Copy the entire
    \texttt{<answer>...</answer>} block verbatim from the
    input to your output. Do not rewrite, reformat,
    re-space, summarize, paraphrase, or ``clean up'' the
    contents inside the tag. Do not insert any whitespace,
    line breaks, commas, dashes, dots, or quotes inside the
    tag. Do not change the tag spelling or casing.

    \item The 36-character string inside
    \texttt{<answer>...</answer>} must remain exactly
    36 characters, with exactly the same digits in exactly
    the same order.

    \item Place the \texttt{<answer>...</answer>} block at
    the very end of your output, on its own, with nothing
    after it.

    \item If the input does not contain a well-formed
    \texttt{<answer>...</answer>} block, still produce your
    output ending with a single
    \texttt{<answer>...</answer>} block whose contents are
    copied verbatim from whatever final answer string the
    input provided, without any modification.
\end{itemize}

Do not turn the response into answer-only or summary-only
form. The rewritten text should still read like a Sudoku
solution process leading up to the final
\texttt{<answer>...</answer>} tag, not just a conclusion.

Output only the paraphrased text, ending with the unchanged
\texttt{<answer>...</answer>} block.

\end{promptbox}

\end{minipage}
\caption{Sudoku-$6\times6$ prompts. Left: hint-generation prompts and the rephrasing user prompt. Right: the rephrasing system prompt.}
\label{tab:prompt_templates}
\end{table*}

\subsection{Representative Hint Examples}
\label{app:hint_examples}

Tables~\ref{tab:sudoku_hint_demo}--\ref{tab:dapo_math_hint_demo} give abridged examples of the teacher-only reference context. The displayed answers are preserved from the examples.

Sudoku and Calcudoku coordinates are one-based; Arrow Maze coordinates are zero-based. Only selected Calcudoku cages are displayed, so that panel is an illustration rather than a complete puzzle specification.

\begin{table*}[!t]
\centering
\setlength{\tabcolsep}{6pt}
\begin{tabularx}{\textwidth}{p{0.30\textwidth}X}
\toprule
\textbf{Problem} & \textbf{Verifier-Accepted Hint (Abridged)} \\
\midrule
\textbf{Sudoku-$6\times6$}

\[
\begin{array}{cccccc}
\cdot&1&6&2&\cdot&\cdot\\
\cdot&\cdot&2&\cdot&1&3\\
\cdot&\cdot&\cdot&\cdot&\cdot&\cdot\\
\cdot&\cdot&1&\cdot&4&6\\
5&\cdot&\cdot&\cdot&6&1\\
1&6&\cdot&3&\cdot&5
\end{array}
\]
&
The puzzle can be solved by standard elimination using row, column, and $2\times3$ block constraints.

\textbf{Key deductions:} (1) Row 1 gives $(1,5)=5$, $(1,6)=4$, and $(1,1)=3$. (2) Column 1 and the upper-left block force $(2,1)=4$; row 2 then gives $(2,2)=5$ and $(2,4)=6$. (3) Column 4 and row 5 force $(5,4)=4$; row 5 then gives $(5,3)=3$ and $(5,2)=2$.

\emph{[Several intermediate deductions are omitted.]}

\textbf{Final solution:}
\[
\begin{array}{cccccc}
3&1&6&2&5&4\\
4&5&2&6&1&3\\
6&4&5&1&3&2\\
2&3&1&5&4&6\\
5&2&3&4&6&1\\
1&6&4&3&2&5
\end{array}
\]
\\
\bottomrule
\end{tabularx}
\caption{A representative verifier-accepted hint for Sudoku-$6\times6$.}
\label{tab:sudoku_hint_demo}
\end{table*}

\begin{table*}[!t]
\centering
\setlength{\tabcolsep}{6pt}
\begin{tabularx}{\textwidth}{p{0.43\textwidth}X}
\toprule
\textbf{Problem} & \textbf{Verifier-Accepted Hint (Abridged)} \\
\midrule
\textbf{Calcudoku-$5\times5$}

Selected cage constraints:
\[
\begin{array}{ll}
(3,3),(4,1),(4,2),(4,3) &: 30\times\\
(3,5),(4,5),(5,5) &: 12+\\
(5,1),(5,2),(5,3),(5,4) &: 11+\\
(1,4),(2,4),(3,4),(4,4) &: 13+\\
(1,3),(2,3) &: 4\div
\end{array}
\]
&
\textbf{Key deductions:} (1) The $4\div$ cage forces $(1,3)$ and $(2,3)$ to contain $\{1,4\}$. (2) The $11+$ cage in row 5 forces its four cells to contain $\{1,2,3,5\}$, yielding $(5,5)=4$. (3) The $12+$ cage then forces $(3,5)$ and $(4,5)$ to contain $\{3,5\}$. (4) The $13+$ cage forces its four cells to contain $\{1,3,4,5\}$. Column 4 therefore gives $(5,4)=2$, and continued constraint propagation gives $(4,4)=4$.

\emph{[Additional row, column, and cage deductions are omitted.]}

\textbf{Final solution:}
\[
\begin{array}{ccccc}
5&4&1&3&2\\
2&3&4&5&1\\
4&2&5&1&3\\
3&1&2&4&5\\
1&5&3&2&4
\end{array}
\]
\\
\bottomrule
\end{tabularx}
\caption{A representative verifier-accepted hint for Calcudoku-$5\times5$.}
\label{tab:calcudoku_hint_demo}
\end{table*}

\begin{table*}[!t]
\centering
\setlength{\tabcolsep}{6pt}
\begin{tabularx}{\textwidth}{p{0.34\textwidth}X}
\toprule
\textbf{Problem} & \textbf{Verifier-Accepted Hint (Abridged)} \\
\midrule
\textbf{Arrow Maze-$6\times6$}

\[
\begin{array}{cccccc}
1&X&3&\uparrow&1&X\\
X&X&\downarrow&1&\leftarrow&1\\
1&\nwarrow&\downarrow&1&X&1\\
X&\leftarrow&3&\downarrow&X&1\\
\leftarrow&1&\leftarrow&1&X&1\\
\leftarrow&1&\leftarrow&1&\leftarrow&1
\end{array}
\]
&
\textbf{Key deductions:} (1) $(0,5)=\rightarrow$: the numbered cell at $(0,4)$ has no other available direction, so its single arrow points right. (2) $(0,1)=\leftarrow$: the 3 at $(0,2)$ already has a downward ray of length two and therefore requires one additional arrow. (3) $(1,1)=\searrow$: the 1 at $(0,0)$ cannot extend to the right because $(0,1)$ has already been assigned to the ray originating from $(0,2)$.

\emph{[Additional deductions and the final constraint check are omitted.]}

\textbf{Final solution:}
\[
\begin{array}{cccccc}
1&\leftarrow&3&\uparrow&1&\rightarrow\\
\uparrow&\searrow&\downarrow&1&\leftarrow&1\\
1&\nwarrow&\downarrow&1&\leftarrow&1\\
\leftarrow&\leftarrow&3&\downarrow&\leftarrow&1\\
\leftarrow&1&\leftarrow&1&\leftarrow&1\\
\leftarrow&1&\leftarrow&1&\leftarrow&1
\end{array}
\]
\\
\bottomrule
\end{tabularx}
\caption{A representative verifier-accepted hint for Arrow Maze-$6\times6$.}
\label{tab:arrow_hint_demo}
\end{table*}

\begin{table*}[!t]
\centering
\setlength{\tabcolsep}{6pt}
\begin{tabularx}{\textwidth}{p{0.40\textwidth}X}
\toprule
\textbf{Problem} & \textbf{Verifier-Accepted Hint (Abridged)} \\
\midrule
\textbf{DAPO-Math}

Given that nonzero real numbers $x$ and $y$ satisfy
\[
x+\frac{1}{y}=3, \qquad y+\frac{1}{x}=4,
\]
find the value of $xy+\frac{1}{xy}$.
&
Use the product $p=xy$ to connect the two equations.

\textbf{Key deductions:} (1) Clearing the denominators reveals the shared relation $p+1=3y=4x$. (2) Combining these relations gives a constraint involving only $p$. (3) This constraint fixes the requested expression $p+1/p$, regardless of which valid root is taken.

\emph{[Intermediate algebra is omitted.]}

\textbf{Final answer:} $10$.
\\
\bottomrule
\end{tabularx}
\caption{A representative verifier-accepted hint for DAPO-Math.}
\label{tab:dapo_math_hint_demo}
\end{table*}

\section{Local Analysis of Hybrid Updates}
\label{app:local_derivations}

\subsection{Notation and Assumptions}
Fix a sampled prefix, token $a$, and group-relative advantage $A$. Let $p=\operatorname{softmax}(z)$ be the current student distribution and $q$ the distribution of the frozen initial teacher $\pi_{\theta_0}$ under a specified privileged context. Assume $p(v),q(v)>0$. Define $u=-\nabla_z\ell$, the negative token-loss gradient. Teacher probabilities, detached magnitude coefficients, and $A$ are held fixed when differentiating the student surrogate.

We evaluate the clipped policy-gradient surrogate at the rollout policy ($\rho=1$), with policy clipping inactive. Separate reference regularization and positive normalization factors are omitted. These assumptions concern one optimization surrogate at one prefix, not derivatives through the sampling process or the direction of a full optimizer step. For binary rewards, zero-variance groups use the convention $A=0$, including all-correct and all-incorrect groups.

\subsection{Magnitude Modulation and Its Bounds}
At this local point, the surrogate has the same gradient as $\ell_{\mathrm M}=-A c(q,a)\log p(a)$, where $c$ is the detached bracketed coefficient in Eq.~\ref{eq:magnitude}. Since $\nabla_z\log p(a)=e_a-p$,
\begin{equation}
u_{\mathrm M}=A c(q,a)(e_a-p).
\label{eq:local_magnitude}
\end{equation}
For $0\leq\lambda\leq1$ and $0\leq\epsilon_w<1$,
\begin{equation}
1-\lambda\epsilon_w\leq c(q,a)\leq1+\lambda\epsilon_w.
\label{eq:magnitude_bound}
\end{equation}
Thus, $A>0$ implies $u_{\mathrm M}(a)=A c(q,a)(1-p(a))\geq0$. When $A=0$, the reward-derived surrogate gradient is zero for every teacher context. With the reported $\lambda=1$ and $\epsilon_w=0.2$, the coefficient lies in $[0.8,1.2]$.

This sign statement applies to the sampled-token component of a local logit gradient. Softmax normalization couples all vocabulary coordinates, and parameter sharing couples positions. Consequently, positive coefficients do not guarantee preservation of the direction of the aggregate parameter update.

\subsection{Reverse-KL Distillation}
For $\ell_{\mathrm K}=D_{\mathrm{KL}}(p\|q)=\sum_v p(v)\log[p(v)/q(v)]$, differentiating through softmax gives
\begin{equation}
\frac{\partial\ell_{\mathrm K}}{\partial z_j}
=p(j)\left[\log\frac{p(j)}{q(j)}-D_{\mathrm{KL}}(p\|q)\right].
\label{eq:rkl_derivative}
\end{equation}
Equivalently, with $b_q(v)=\log q(v)-\log p(v)$,
\begin{equation}
u_{\mathrm K}(v)=p(v)\bigl[b_q(v)-\mathbb E_{w\sim p}b_q(w)\bigr].
\label{eq:local_updates}
\end{equation}
For positive distributions, $u_{\mathrm K}=0$ implies that $\log[q(v)/p(v)]$ is constant in $v$; normalization then gives $p=q$. Distinct teacher and student distributions can therefore supply a nonzero local logit gradient even when the group-relative advantage vanishes. This does not guarantee useful guidance or a nonzero aggregate parameter update. Offline hints make such guidance available without requiring a correct sibling; this availability is shared by the hint-conditioned baselines.

These identities concern exact categorical reverse KL. With a top-$K$ partition and a residual bucket, the categorical identity applies to the grouped probabilities, followed by the chain rule to the original logits. A finite sampled-token estimator need not equal the full-vocabulary gradient on each update.

\subsection{Effects of Changing Teacher Context}
\label{app:context_updates}
\label{app:context_decomposition}
Hold the student distribution and scored prefix fixed. Let $q^{\mathrm N}$ and $q^{\mathrm R}$ use the same frozen teacher and correct student response, without and with the rephrasing instruction. For $\delta(v)=\log q^{\mathrm R}(v)-\log q^{\mathrm N}(v)$,
\begin{equation}
\begin{aligned}
u_{\mathrm M}^{\mathrm R}-u_{\mathrm M}^{\mathrm N}
&=A(c_{\mathrm R}-c_{\mathrm N})(e_a-p),\\
u_{\mathrm K}^{\mathrm R}(v)-u_{\mathrm K}^{\mathrm N}(v)
&=p(v)\bigl[\delta(v)\\
&\quad-\mathbb E_{w\sim p}\delta(w)\bigr].
\end{aligned}
\label{eq:context_effect}
\end{equation}
Magnitude modulation rescales a fixed token's local direction, with
\begin{equation}
\begin{aligned}
\|u_{\mathrm M}^{\mathrm R}-u_{\mathrm M}^{\mathrm N}\|
&=|A|\,|c_{\mathrm R}-c_{\mathrm N}|\,\|e_a-p\|\\
&\leq2\lambda\epsilon_w|A|\,\|e_a-p\|.
\end{aligned}
\end{equation}
Reverse KL instead changes centered preferences across the vocabulary and can oppose sampled-token reinforcement. Neither identity establishes that a particular rephrasing instruction improves training.

A teacher--student score difference also contains a policy difference. Let $q_t^0=\pi_{\theta_0}(\cdot\mid x,y_{<t})$ denote the frozen teacher without hindsight context. Then
\begin{equation}
\log\frac{q_t^{\mathrm R}(y_t)}{p_t(y_t)}
=\log\frac{q_t^{\mathrm R}(y_t)}{q_t^0(y_t)}
+\log\frac{q_t^0(y_t)}{p_t(y_t)}.
\label{eq:context_policy_decomposition}
\end{equation}
Only the first term changes context while holding model parameters fixed. Comparing R and N cancels the common student term, but can still reflect lexical, copying, or formatting preferences. Because the complete response is supplied as hindsight, increased token likelihood is not by itself evidence of causal importance.

\subsection{Trajectory Scaling and Token Redistribution}
\label{sec:redistribution}
\label{app:weight_placement}
For one trajectory, let fixed $\alpha_t\geq0$ satisfy $\sum_t\alpha_t=1$, let $g_t=\nabla_\theta\log p_t(y_t)$, and define $\bar c=\sum_t\alpha_t c_t$. Substitution gives
\begin{equation}
\begin{aligned}
G_{\mathrm M}&=\sum_t\alpha_t A c_t g_t\\
&=\bar c\,G_{\mathrm{PG}}
+\sum_t\alpha_t A(c_t-\bar c)g_t,\\
G_{\mathrm{PG}}&=\sum_t\alpha_t A g_t.
\end{aligned}
\label{eq:redistribution}
\end{equation}
The first term scales the trajectory-level gradient; the second redistributes weights across positions. Constant coefficients, including a common clipping saturation value, eliminate the second term. Nonconstant coefficients permit redistribution but do not establish its benefit, since token gradients can also cancel.

For multiple trajectories, nonnegative fixed normalization weights $\alpha_{it}$ yield
\begin{equation}
\begin{aligned}
\|G_{\mathrm M}-G_{\mathrm{PG}}\|
&\leq\sum_{i,t}\alpha_{it}|A_i|\,|c_{it}-1|\,\|g_{it}\|\\
&\leq\lambda\epsilon_w
\sum_{i,t}\alpha_{it}|A_i|\,\|g_{it}\|.
\end{aligned}
\label{eq:aggregate_magnitude_bound}
\end{equation}
This absolute bound inherited from bounded magnitude weights does not imply the same relative bound with respect to $\|G_{\mathrm{PG}}\|$. Adaptive optimization and regularization further separate this gradient sum from an optimizer step.

The GRPO on Success row in Table~\ref{tab:routing_ablation} is the $c_t=1$ control with the failure branch retained. It tests the total contribution of successful-trajectory teacher weighting. Trajectory-wise constant weights $c_t=\bar c$ and within-trajectory permutations would further distinguish scaling from weight placement; these controls are not reported here. Permutation preserves the mean under uniform normalization; nonuniform normalization requires matching the weighted mean as well. Neither control automatically preserves the gradient norm.

\section{Token-Level Rephrasing Analysis}
\label{app:rephrase_token_scores}
\begin{figure*}[!t]
\centering
\includegraphics[width=\textwidth]{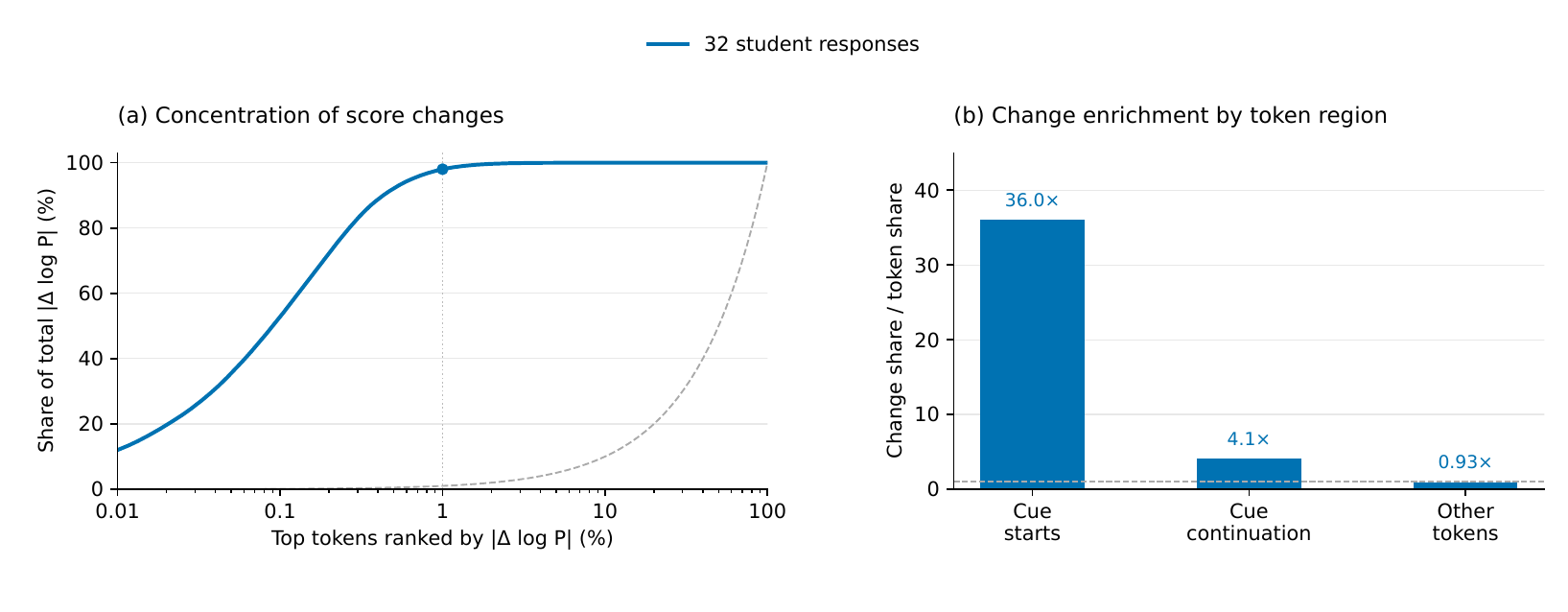}
\caption{Absolute score changes from rephrasing over 32 correct Sudoku-$6\times6$ responses (188,062 tokens). Left: cumulative change share, ranked by magnitude. Right: change share divided by token share for each category.}
\label{fig:score_concentration}
\end{figure*}

Figure~\ref{fig:teacher_student_schematic} illustrates the directions of token-level score changes. Figure~\ref{fig:score_concentration} measures their absolute magnitudes on successful responses. The local update identities in Appendix~\ref{app:local_derivations} distinguish these score changes from their effects on optimization.

\subsection{Paired Scoring Protocol}
Both teacher conditions receive the same problem and complete correct student response; only the instruction to rephrase differs. The frozen teacher scores the unchanged tokens by teacher forcing, without generating a rewritten response. Let $q_t^k=\pi_{\theta_0}(\cdot\mid x,r^k,y_{<t})$, $p_t=\pi_\theta(\cdot\mid x,y_{<t})$, and N/R denote no rephrasing/rephrasing. The main-text schematic depicts
\begin{equation}
\begin{aligned}
\Delta_t^k&=\log q_t^k(y_t)-\log p_t(y_t),\\
k&\in\{\mathrm N,\mathrm R\}.
\end{aligned}
\label{eq:rephrase_score_visualization}
\end{equation}
For a fixed generated token and student score, the instruction-induced change is
\begin{equation}
\begin{aligned}
\delta_t&=\Delta_t^{\mathrm R}-\Delta_t^{\mathrm N}\\
&=\bigl[\log q_t^{\mathrm R}(y_t)-\log p_t(y_t)\bigr]\\
&\quad-\bigl[\log q_t^{\mathrm N}(y_t)-\log p_t(y_t)\bigr]\\
&=\log q_t^{\mathrm R}(y_t)-\log q_t^{\mathrm N}(y_t).
\end{aligned}
\label{eq:rephrase_score_change}
\end{equation}
The student term cancels. Figure~\ref{fig:score_concentration} reports $|\delta_t|$, labeled $|\Delta\log P|$, so its statistic measures adjustment magnitude rather than direction.

\subsection{Concentration of Absolute Score Changes}
\label{app:score_concentration}
The analysis covers 32 correct Sudoku-$6\times6$ student responses, totaling 188,062 tokens. The left panel ranks all tokens by decreasing $|\delta_t|$. Its logarithmic horizontal axis gives the selected token fraction, and its vertical axis gives their cumulative share of total absolute change. The top 0.1\% of tokens contribute approximately 52.9\% of the total; the top 1\% contribute approximately 98.0\%. The gray dashed curve is the reference for uniformly distributed changes.

The right panel separates three categories. \emph{Cue starts} are the first tokens produced by the tokenizer for matched planning or checking phrases, such as ``Let's label\ldots'' or ``But wait\ldots''. \emph{Cue continuation} comprises the remaining tokens inside these phrases, and \emph{other tokens} comprises all remaining positions. Enrichment is a category's absolute-change share divided by its token share, equivalently its mean $|\delta_t|$ divided by the global mean. The enrichments are $36.0\times$, $4.1\times$, and $0.93\times$, respectively. Cue starts account for 0.175\% of tokens and 6.30\% of total absolute change.

Thus, rephrasing does not broadly alter teacher scores across the trajectory. Instead, its effect is highly concentrated on a small subset of token positions, particularly at planning and checking cue starts. This concentration suggests that the benefit of rephrasing is mediated primarily through a small number of high-impact decision points, where changes in teacher scores induce disproportionately large changes in the token-level update magnitudes.

\subsection{Relation to the Training Updates}
In Figure~\ref{fig:teacher_student_schematic} green/red backgrounds indicate higher/lower teacher probabilities than the student's. Red outlines mark the reasoning setup ``Let's label''; green outlines mark the potentially redundant self-interruption ``But wait''. The illustrated changes suggest stronger support for starting the solution process and weaker support for the interruption.

For a correct response in a mixed-reward group, $A>0$ and $w_t=\exp(\Delta_t)$. The coefficient in Eq.~\ref{eq:magnitude} is therefore nondecreasing in the teacher--student score difference. Higher or lower teacher scores can strengthen or weaken positive reinforcement without reversing the advantage sign, although clipping may reduce or eliminate these differences. In an all-correct group, $A=0$, so the reward-derived update remains zero.

Taken together with the concentration analysis above, these results suggest a sparse mechanism for successful-trajectory supervision: rephrasing changes teacher scores at only a small subset of token positions, and these changes induce disproportionately large differences in their update magnitudes. By selectively strengthening or weakening reward-derived updates at these high-impact positions, rephrasing provides a focused form of token-level credit assignment that can improve reasoning performance.

\subsection{Successful-Trajectory Context Ablation}
\label{app:success_context}

We further examine the choice of privileged context
for successful trajectories. We compare conditioning
the teacher on an external reference hint with
conditioning it on the student's verified response,
while disabling rephrasing and keeping the failure
branch unchanged in both settings.

\par
\vspace{\intextsep}
\noindent\begin{minipage}{\columnwidth}
\captionsetup{type=table}
\centering
\small
\setlength{\tabcolsep}{6pt}
\begin{tabular}{lcc}
\toprule
Successful-Trajectory Context & Sudoku & Calcudoku \\
\midrule
External reference hint & 61.00 & 41.33 \\
Verified student response & \textbf{61.50} & \textbf{43.17} \\
\bottomrule
\end{tabular}
\caption{Validation accuracy (\%) under different
successful-trajectory teacher contexts without
rephrasing. Both settings retain hint-conditioned
reverse KL for failed trajectories.}
\label{tab:success_context}
\end{minipage}
\par\vspace{\intextsep}

As shown in Table~\ref{tab:success_context}, conditioning on the verified student response outperforms using an external reference hint on Sudoku and Calcudoku. Since many reasoning problems such as Calcudoku may admit multiple valid reasoning paths, external hints may differ from the student's correct trajectory. Conditioning on the verified response avoids this potential mismatch, with rephrasing providing further gains as discussed in Section~\ref{sec:rephrasing}.

\twocolumn[{
\begin{minipage}{\textwidth}
\captionsetup{type=figure}
\centering
\includegraphics[width=0.74\textwidth]{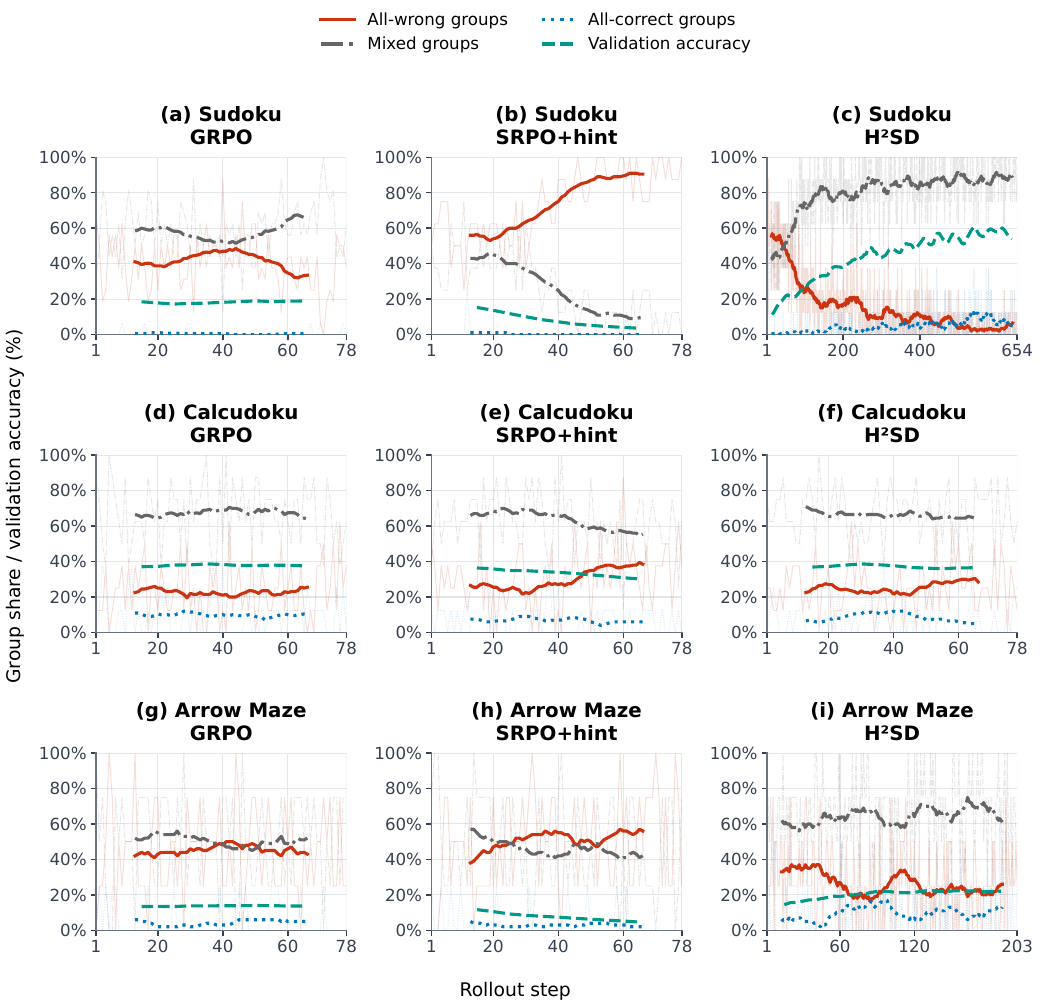}
\caption{Training group composition and greedy validation accuracy on Sudoku, Calcudoku, and Arrow Maze. Training horizons differ because GRPO and SRPO+Hint converged earlier, while H$^2$SD continued to improve on some tasks.}
\label{fig:group_grpo}
\end{minipage}
\par\vspace{\dbltextfloatsep}
}]
\section{Additional Training Diagnostics}
\label{app:training_diagnostics}

\subsection{Group Composition and Routing Dynamics}
\label{app:group_routing}

Figure~\ref{fig:group_grpo} compares the evolution of training group composition and validation accuracy for GRPO, SRPO+Hint and H$^2$SD. All-wrong, mixed, and all-correct fractions use rollout groups as their denominator and sum to one. With fixed group size and binary rewards, all-wrong and all-correct groups have zero group-relative advantage. Hint-conditioned distillation can still supervise failed trajectories in all-wrong groups, whereas reward-based updates vanish in such groups.

Training horizons differ across methods because runs were stopped after validation performance had plateaued over the observed training horizon. The validation curves report greedy accuracy throughout
training, while Table~\ref{tab:main_results} reports
the performance of the selected checkpoints under the
evaluation protocol described in
Appendix~\ref{app:eval_protocol}. H$^2$SD is trained for longer on Sudoku and Arrow Maze because its validation accuracy continues to improve beyond this point. The different horizontal ranges therefore reflect differences in convergence behavior rather than a common fixed step budget.

On Sudoku, H$^2$SD progressively reduces the fraction of all-wrong groups as validation accuracy increases, whereas SRPO+Hint moves toward predominantly all-wrong groups while its validation accuracy declines. Calcudoku is dominated by mixed groups for all three methods, with relatively stable validation accuracy. On Arrow Maze, H$^2$SD maintains a substantially lower all-wrong fraction than SRPO+Hint while achieving higher validation accuracy.

\twocolumn[{
\begin{minipage}{\textwidth}
\captionsetup{type=figure}
\centering
\includegraphics[width=0.83\textwidth]{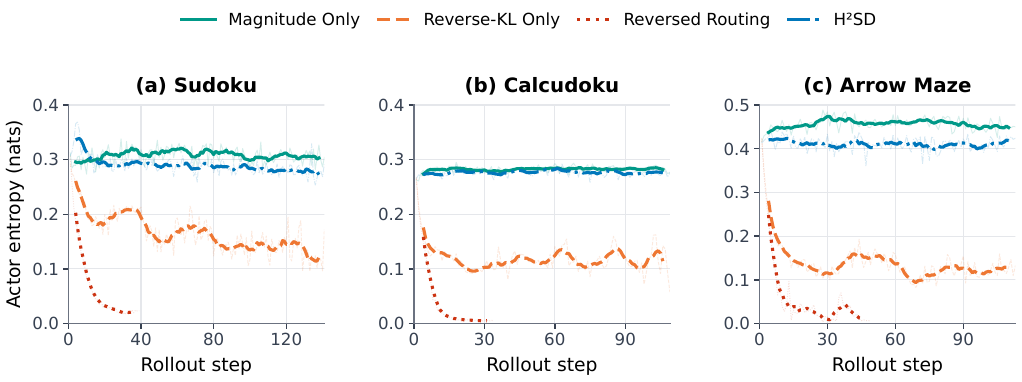}
\caption{Actor entropy under four update strategies on Sudoku, Calcudoku, and Arrow Maze. Light curves show recorded values; dark curves are centered means.}
\label{fig:update_entropy_all}
\end{minipage}
\par\bigskip
\begin{minipage}{\textwidth}
\captionsetup{type=figure}
\centering
\includegraphics[width=0.76\textwidth]{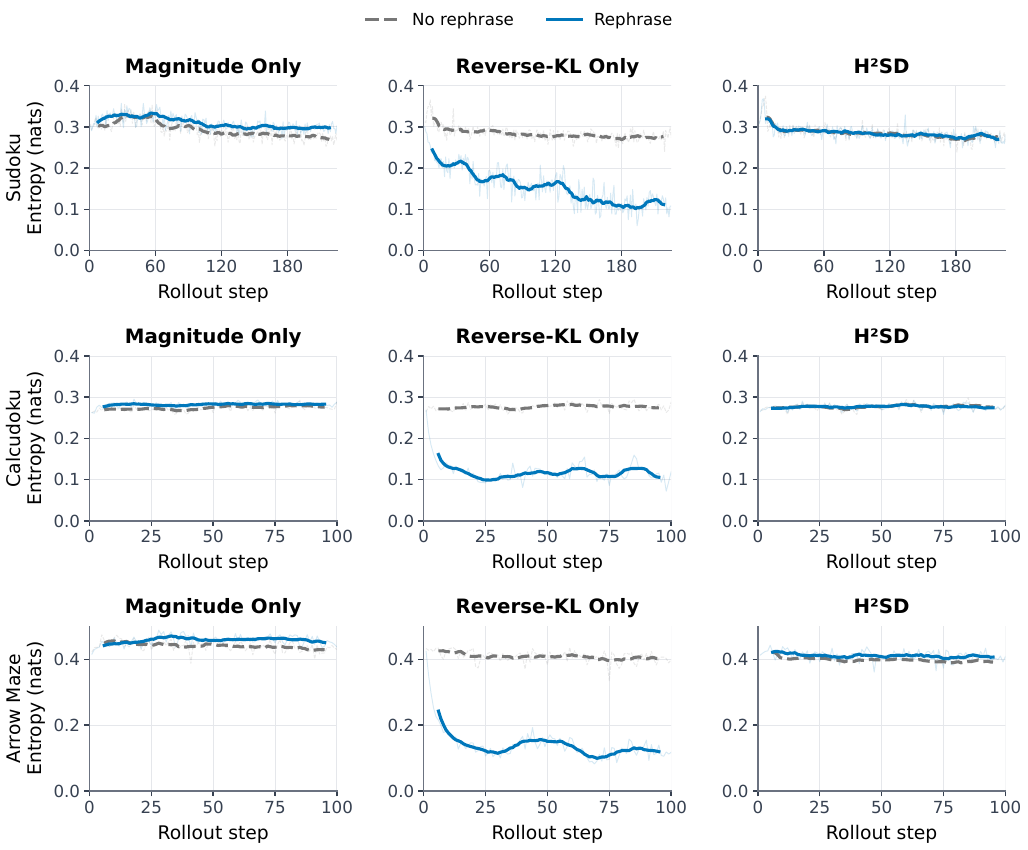}
\caption{Rephrasing diagnostics across logic tasks (rows) and update strategies (columns). Dark curves use centered means.}
\label{fig:rephrase_entropy_all}
\end{minipage}
\par\vspace{\dbltextfloatsep}
}]
\subsection{Actor Entropy under Routing and Rephrasing}
\label{app:entropy_analysis}
Figures~\ref{fig:update_entropy_all} and~\ref{fig:rephrase_entropy_all} report Qwen3-30B-A3B-Instruct actor entropy on logic-task training rollouts. The plotted Reverse-KL Only and Reversed Routing runs exhibit substantially stronger entropy collapse than Magnitude Only and H$^2$SD, together with lower or stagnant accuracy. This pattern suggests that overly aggressive distribution matching can concentrate the policy too quickly, suppressing exploration and contributing to the observed performance degradation. \par\noindent
Rephrasing further intensifies the entropy collapse under Reverse-KL Only, consistent with its reduced accuracy in Table~\ref{tab:rephrasing}. In contrast, Magnitude Only and H$^2$SD maintain higher policy entropy and achieve more stable performance, supporting the view that preserving sufficient exploration is important for effective self-distillation.

\twocolumn[{
\begin{minipage}{\textwidth}
\captionsetup{type=figure}
\centering
\includegraphics[width=0.82\textwidth]{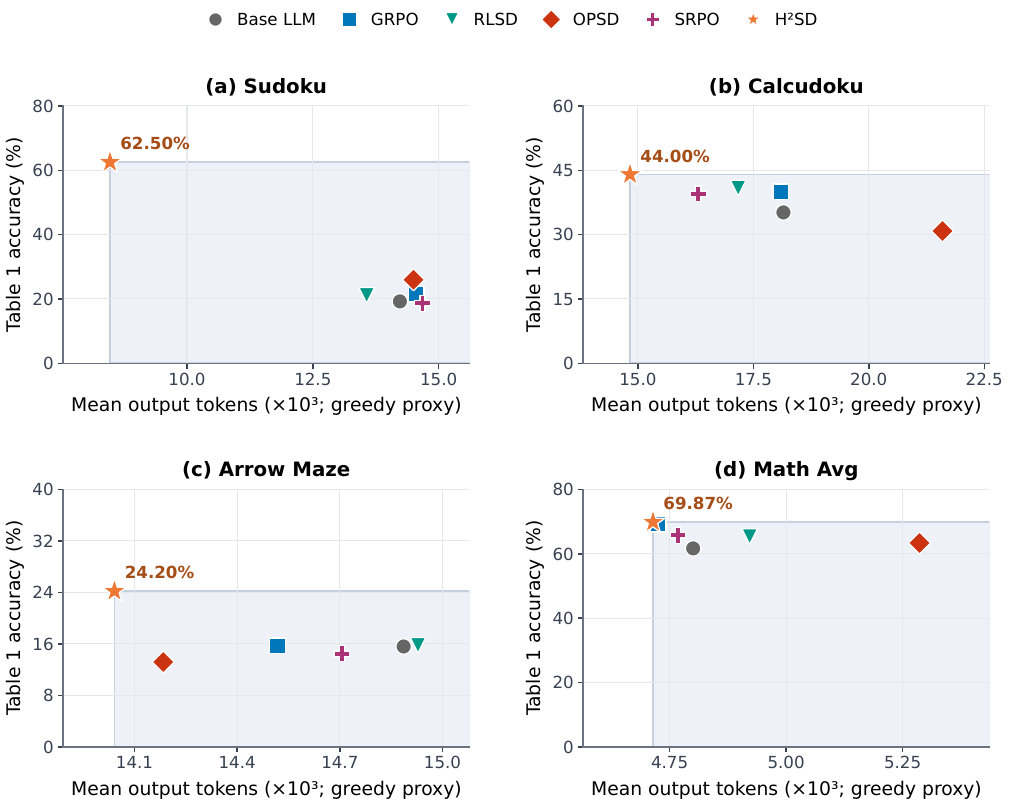}
\caption{Table~\ref{tab:main_results} accuracy versus archived greedy output length for Qwen3-30B-A3B-Instruct. Shading marks coordinates with no shorter outputs and no higher accuracy than H$^2$SD. Accuracy and length use separate evaluations.}
\label{fig:length_tradeoff_all}
\end{minipage}
\par\vspace{\dbltextfloatsep}
}]
\subsection{Accuracy and Generation-Cost Diagnostics}
\label{app:length_analysis}

Figure~\ref{fig:length_tradeoff_all} pairs the evaluation accuracy reported in Table~\ref{tab:main_results} with retokenized lengths from archived single-response greedy evaluations. RLSD and SRPO both use hints. H$^2$SD has the highest accuracy and lowest archived length in all four panels. On Sudoku, its 62.50\% accuracy and approximately 8,470 tokens compare with GRPO's 21.50\% and 14,552 tokens. The Math length difference is smaller, with accuracies of 69.87\% and 69.18\%.

For a matched evaluation under Eq.~\ref{eq:eval_accuracy}, define
$a=\widehat{\mathrm{Acc}}_S/100$. When the conditional means exist,
\par
\begin{equation}
\begin{aligned}
\bar L
&=\frac{1}{NS}\sum_{i=1}^{N}
\sum_{s=1}^{S}|y_{i,s}|\\
&=a\bar L_{\mathrm{correct}}
+(1-a)\bar L_{\mathrm{incorrect}}.
\end{aligned}
\label{eq:length_decomposition}
\end{equation}

Longer failures can make higher accuracy alone lower $\bar L$; fewer repeated checks could instead lower $\bar L_{\mathrm{correct}}$. Testing the mechanism suggested by Figure~\ref{fig:teacher_student_schematic} requires matched conditional lengths and repeated-check frequencies, together with the weight-placement controls in Appendix~\ref{app:weight_placement}. Output length excludes hint construction, teacher scoring, and training compute.



\end{document}